\documentclass{amos}
\usepackage{bm}
\usepackage{mathrsfs}
\usepackage{amsmath}
\usepackage[colorlinks=true, pdfstartview=FitV, linkcolor=black, citecolor= black, urlcolor= black]{hyperref}
\usepackage{footnpag}			      	
\usepackage{url}
\usepackage[ruled, vlined, linesnumbered]{algorithm2e}
\usepackage{amssymb, graphicx, mathtools, soul, xcolor, float, algorithmic, siunitx, booktabs}
\usepackage{multirow,cleveref}
\usepackage{booktabs}
\usepackage{subcaption}
\usepackage{threeparttable}
\usepackage[sort, numbers]{natbib}
\usepackage{gensymb}
\usepackage{enumitem}
\usepackage{hyperref}

\title{\TitleFont DreamSat-Bench: Development and Initial Testing of a Testbed for AI-Based Pose Estimation from 3D Reconstruction}

\author{
Alex Posadas-Nava\footnote{Graduate Student, Department of Aeronautics and Astronautics, \textit{aposadas@mit.edu}}, \;
August Berne\footnote{Undergraduate Student, Department of Aeronautics and Astronautics, \textit{adberne@mit.edu}}, \;
Giovanni Lavezzi\footnote{Research Scientist, Department of Aeronautics and Astronautics, \textit{glavezzi@mit.edu}}, \;
Kareena Shah\footnote{Undergraduate Student, Department of Aeronautics and Astronautics, \textit{kareenas@mit.edu}}, \;
Alejandro Carrasco\footnote{Graduate Student, Department of Aeronautics and Astronautics, \textit{acarra@mit.edu}}, \;
Josiane Uwumukiza\footnote{Intern, Department of Aeronautics and Astronautics, \textit{josianeu@mit.edu}}
\\ Massachusetts Institute of Technology 

\and \vspace{2 pt}

Giacomo Battaglia\footnote{Ph.D. Student, Department of Aerospace Science and Technology, \textit{giacomo.battaglia@polimi.it}}, \;
Paolo Panicucci\footnote{Assistant Professor, Department of Aerospace Science and Technology, \textit{paolo.panicucci@polimi.it}}
\\ Politecnico di Milano 

\and \vspace{-7 pt}

Minduli C. Wijayatunga\footnote{Assistant Professor, Department of Aerospace Engineering, \textit{minduli@illinois.edu}}
\\ University of Illinois Urbana-Champaign

\and \vspace{-7 pt}

Victor Rodriguez-Fernandez\footnote{Associate Professor, Department of Computer Systems Engineering, \textit{victor.rfernandez@upm.es}}, \;
\\ Universidad Politécnica de Madrid 

\and 
\vspace{-7 pt}

Richard Linares\thanks{Associate Professor, Department of Aeronautics and Astronautics, \textit{linaresr@mit.edu}} 
\\ Massachusetts Institute of Technology
}

\date{}

\makeatletter
\def\@fnsymbol#1{\ensuremath{\number#1}}

\def\blfootnote{\xdef\@thefnmark{}\@footnotetext}
\makeatother

\begin{document} 

\maketitle

\begin{abstract}
\normalsize
This paper presents the development and initial testing of DreamSat-Bench, a modular rendezvous and proximity operation testbed designed to benchmark AI-based relative navigation techniques. By integrating a software- and hardware-in-the-loop robotic pipeline, the platform enables a seamless transition from digital simulation to physical reality. DreamSat-Bench unifies state-of-the-art robotic learning tools such as MuJoCo, Isaac Lab, and LeRobot into a single benchmarking platform, utilizing robotic arms to trace 3D trajectories. The platform allows for extensive customization of orbital environments and lighting to evaluate the simulation-to-reality gap. We demonstrate the testbed's utility by evaluating an end-to-end vision-based navigation pipeline that pairs DreamSat, a generative AI framework for single-view 3D reconstruction, with FoundationPose for zero-shot 6-DoF tracking of unseen spacecraft. Initial testing explores mission-representative orbital segments, including fixed-point station-keeping and fly-around characterization. Through a series of parametric studies, we quantify the impacts of reconstruction latency, mesh resolution, orbital range, and illumination geometry on pose estimation accuracy. Finally, a preliminary hardware-in-the-loop campaign qualitatively validates the physical deployment of the pipeline, identifying target symmetry and accumulated tracking drift as critical factors for robust navigation. DreamSat-Bench provides a rigorous framework for maturing autonomous navigation with unprepared space assets in the absence of prior geometric models.
\end{abstract}

\blfootnote{The authors A. Posadas-Nava and A. Berne have equally contributed to this work.}

\blfootnote{\scriptsize Research was sponsored by the Department of the Air Force Artificial Intelligence Accelerator and was accomplished under Cooperative Agreement Number FA8750-19-2-1000. The views and conclusions contained in this document are those of the authors and should not be interpreted as representing the official policies, either expressed or implied, of the Department of the Air Force or the U.S. Government. The U.S. Government is authorized to reproduce and distribute reprints for Government purposes notwithstanding any copyright notation herein. DO NOT USE ANY MARKINGS THAT APPEAR BELOW THIS LINE.}

\section{Introduction}
As the density of resident space objects (RSOs) in Low Earth Orbit continues to grow, the requirements for Space Situational Awareness (SSA) and Space Domain Awareness (SDA) have shifted from simple tracking to complex characterization. Future missions involving Active Debris Removal (ADR) and on-orbit servicing require the ability to perform Rendezvous and Proximity Operations (RPO) with unprepared or non-cooperative objects, targets for which no prior 3D model or cooperative markers exist. 

Although software-only simulations provide scalability, physical validation is essential to address the simulation-to-reality gap. To this end, several world-class facilities have been established, such as the Testbed for Rendezvous and Optical Navigation (TRON) at Stanford SLAB, which has produced pose estimation datasets like SPEED+ \cite{Park2022SPEEDplus}, and the European Space Agency’s (ESA) Rendezvous, Approach and Landing Simulator (GRALS) facility used for validating Guidance, Navigation, and Control (GNC) algorithms on scaled mockups \cite{pasqualetto2021ground}. Other researchers have utilized planar air-bearing testbeds for contact dynamics and adaptive control, including Carleton's Spacecraft Proximity Operations Testbed (SPOT) \cite{carleton1, carleton2, carleton3} and Caltech's Center for Autonomous Systems and Technologies (CAST) Multi-Spacecraft Testbed for Autonomy Research (M-STAR) facility \cite{nakka2018}. Furthermore, laboratories like Florida Tech's Orbital Robotic Interaction, On-Orbit Servicing and Navigation (ORION) \cite{caruso2023} and KTH's Space Robotics Lab \cite{roque25} have expanded these capabilities into the realms of neural 3D reconstruction and modular, multi-agent autonomy. Notably, the Proximity Operation of Spacecraft: Experimental hardware-In-the-loop DYNamic Simulator (POSEIDYN) testbed at the Naval Postgraduate School \cite{romano2007} has established a long-standing baseline for validating planar proximity maneuvers and robotic interaction with non-cooperative targets using floating simulators. 
Recent efforts have focused on bridging high-fidelity astrodynamics frameworks like Basilisk \cite{basilisk} with modular robotics middleware such as Robot Operating System (ROS) 2 to enable a unified autonomy pipeline and seamless transitions between orbital simulation and hardware-in-the-loop (HITL) testing \cite{krantz2025}.
Comprehensive validation frameworks have been proposed that utilize a multi-layered approach, employing software-based simulation for high-level mission autonomy, arm-mounted optical testbeds for verifying perception algorithms with physical scale models, and planar air-bearing tables to emulate the contact-rich dynamics of robotic docking and manipulation, to simulate in-orbit servicing scenarios \cite{Minduli_cortex_testbed}.
End-to-end frameworks like CORTEX have successfully integrated convex-based trajectory optimization with modular hardware-in-the-loop testbeds to demonstrate safe and precise guidance for close-range rendezvous and docking maneuvers \cite{cortex}.

Spacecraft pose estimation is progressing from direct convolutional neural network (CNN) regression \cite{sharma2018pose} and the identification of mission-specific structural markers \cite{candan2025markers} to zero-shot primitive-based abstractions \cite{park2024improving} that utilize diverse datasets like SPE3R \cite{spe3r_data} to generalize across unseen spacecraft geometries. These methods have been further enhanced by ambiguity-aware loss functions to mitigate shape-pose inconsistencies in symmetric targets \cite{bates2025removing}. To move beyond these geometric primitives, recent work has adopted 3D Gaussian Splatting (3DGS) for high-fidelity reconstruction \cite{nguyen2024}, specifically leveraging primitive-based initializations to accelerate convergence and provide a geometric prior for unknown targets \cite{hucfast}. However, while these abstraction-based methods generalize to novel classes, they often produce coarse models that require high-quality initialization to avoid local minima during high-fidelity refinement. To address this, foundation models and generative synthesis are of great interest. NVIDIA’s FoundationPose (FP) \cite{wen2024foundationpose} provides a unified framework for six-degree-of-freedom (6-DoF) pose estimation and tracking that generalizes to novel objects by leveraging large-scale synthetic data. Generative frameworks like DreamSat \cite{dreamsat, dreamsat2} utilize diffusion-based priors to reconstruct 3D satellite geometries from single-view monocular observations. Building on these reconstructions, DreamSat-Pose \cite{dreamsat-pose} enables single-shot 6-DoF pose estimation of unknown spacecraft by matching dense 2D image features with 3D geometric descriptors through a dual-stream transformer\cite{dreamsat-pose}. Together, these advancements move spacecraft navigation away from rigid geometric priors toward flexible, high-fidelity architectures capable of characterizing non-cooperative targets in real-time.

This paper presents the preliminary development and initial testing of an RPO simulator testbed, DreamSat-Bench, that integrates generative AI-based 3D reconstruction with vision-based pose estimation to enable autonomous proximity maneuvers. 
The RPO simulator testbed, DreamSat-Bench, is a simulation suite designed to test and evaluate physical AI algorithms. DreamSat-Bench unifies state-of-the-art tools like MuJoCo \cite{todorov2012mujoco}, MuJoCo Menagerie \cite{Zakka_MuJoCo_Menagerie_A_2022}, Isaac Lab \cite{nvidia2025isaaclabg}, and LeRobot \cite{cadene2026lerobot} into a single, cohesive benchmarking platform for robotic learning. MuJoCo excels at articulated, contact-rich robotic simulation, which allows us to create RPO mission scenarios using robotic arms. We attach spacecraft models to the robots’ end effectors to trace out trajectories in 3D space. 
The DreamSat-Bench environment enables the emulation of RPO scenarios on physical hardware, effectively facilitating Software-in-the-loop (SITL) and Hardware-in-the-loop (HITL) simulations. 
We leverage the Relative Orbital Elements (ROE) \cite{roe} to design a suite of mission-representative relative motion paths; while the platform is capable of simulating full-proximity approaches, this initial study focuses on characterizing navigation performance during fixed-point observations and fly-around trajectories. These trajectories are then converted into the Hill-Clohessy-Wiltshire (HCW) \cite{hcw_eqs} frame to provide precise Cartesian relative states, which serve as the ground truth (GT) for our camera pointing and simulation logic. The 6-DoF pose estimation pipeline is built upon DreamSat \cite{dreamsat, dreamsat2}, a generative AI framework capable of reconstructing the 3D geometry of a target from a single monocular observation. While this architecture is designed to eventually utilize DreamSat-Pose \cite{dreamsat-pose}, a concurrent effort focused on learned 2D-3D feature matching, this study adopts NVIDIA’s FoundationPose \cite{wen2024foundationpose} as a state-of-the-art representative baseline for pose tracking. This allows for a rigorous evaluation of the testbed’s utility while the specialized DreamSat-Pose pipeline undergoes further refinement.

For the initial testing of DreamSat-Bench, we evaluate the end-to-end pipeline across a suite of mission-representative orbital segments, focusing on fixed-point observations and fly-around characterizations. In SITL simulations, the framework is quantitatively benchmarked by comparing FP estimates against simulation GT across varying reconstruction times, mesh resolutions, orbital ranges, and illumination geometries. These assessments characterize the fundamental trade-offs between computational latency and tracking precision. Finally, a preliminary HITL campaign qualitatively validates the physical deployment of the pipeline, identifying the impact of target symmetry and accumulated tracking drift as key factors for future autonomous navigation refinement.

\section{Fundamentals}
In this section, fundamentals related to relative motion paths, 3D reconstruction, and pose estimation are introduced.

\subsection{Relative Motion Paths}
To design a suite of mission-representative relative motion paths, including fixed-point observations and fly-around characterizations, we leverage the ROE framework \cite{roe}. This parameterization is particularly suited for near-circular orbits as it provides a geometrically intuitive representation of the relative trajectory while avoiding mathematical singularities. The state of the deputy (or chaser) satellite relative to the chief (or target) is defined by the dimensionless ROE vector $\delta \bm{\alpha}$ \cite{sasaki2026predictive}:
\begin{equation}
\delta \bm{\alpha} = 
\begin{bmatrix}
\delta a \\ \delta \lambda \\ \delta e_x \\ \delta e_y \\ \delta i_x \\ \delta i_y
\end{bmatrix} = 
\begin{bmatrix}
(a_d - a)/a \\
(u_d - u) + (\Omega_d - \Omega) \cos i \\
e_d \cos \omega_d - e \cos \omega \\
e_d \sin \omega_d - e \sin \omega \\
i_d - i \\
(\Omega_d - \Omega) \sin i
\end{bmatrix} ,
\label{eq:roe_def}
\end{equation}
where the subscripts $d$ and no subscript denote the deputy and chief satellites, respectively. Here, $a$ is the semi-major axis, $i$ the inclination, $\Omega$ the right ascension of the ascending node (RAAN), and $u$ the mean argument of latitude. The relative geometry is further characterized by the eccentricity vector $\delta \bm{e} = [\delta e_x, \delta e_y]^T$ and the inclination vector $\delta \bm{i} = [\delta i_x, \delta i_y]^T$, which can be expressed in polar form to describe the amplitude and phase of the relative motion:
\begin{equation}
\delta \bm{e} = \delta e \begin{bmatrix} \cos \phi \\ \sin \phi \end{bmatrix}, \quad 
\delta \bm{i} = \delta i \begin{bmatrix} \cos \theta \\ \sin \theta \end{bmatrix} ,
\label{eq:vectors_polar}
\end{equation}
where $\delta e$ and $\delta i$ represent the relative orbit's size, while the phases $\phi$ (relative perigee) and $\theta$ (relative ascending node) determine the orientation of the relative ellipse.
The Cartesian relative states are retrieved by converting the ROEs into the HCW frame. For a chief in a near-circular orbit, the first-order mapping between the ROEs and the Cartesian relative position ($x$: radial, $y$: along-track, $z$: cross-track) as a function of the argument of latitude $u$ is:

\begin{equation}
\begin{aligned}
    x(u) &= a (\delta a - \delta e_x \cos u - \delta e_y \sin u) \\
    y(u) &= a (\delta \lambda - \frac{3}{2} \delta a (u - u_0) + 2\delta e_x \sin u - 2\delta e_y \cos u) \\
    z(u) &= a (\delta i_x \sin u - \delta i_y \cos u)
\end{aligned} \quad .
\label{eq:roe_to_hill}
\end{equation}

where $u_0$ is the initial argument of latitude (set to zero at trajectory initialisation). This mapping constitutes the closed-form solution to the HCW equations of relative motion \cite{hcw_eqs}:

\begin{equation}
\begin{aligned}
\ddot{x} &= 3n^2 x + 2n \dot{y} \\
\ddot{y} &= - 2n \dot{x} \\
\ddot{z} &= - n^2 z 
\end{aligned} \quad ,
\label{eq:hcw_equations}
\end{equation}

where $n = \sqrt{\mu / a^3}$ is the mean motion of the chief orbit and $\mu$ is the standard gravitational parameter of the central body. Rather than integrating \eqref{eq:hcw_equations} numerically, the simulation evaluated \eqref{eq:roe_to_hill} directly over a discrete set of argument-of-latitude values $u \in [0, 2\pi)$, producing a sequence of waypoints that the deputy spacecraft tracks. This approach is exact under the linearization assumptions (near-circular chief orbit, small relative separations) and avoids accumulation of numerical integration error over the orbital period.
Within the MuJoCo simulation environment, coordinates of target and chaser are expressed in the Radial-Transverse-Normal (RTN) Frame. This is the coordinate frame in which the radial direction is oriented directly out of the center of the Earth's simulated location. The Transverse direction is in the direction of the target satellites simulated motion. The Normal is orthogonal to both in a right-handed sense, consequently pointing in the direction of the spacecraft's angular momentum about the Earth.


\subsection{3D Reconstruction Models}
DreamSat \cite{dreamsat, dreamsat2} enables the generation of novel viewpoints from a single scene, which is particularly valuable in the space domain for synthesizing 3D models of unprepared assets (e.g., derelict satellites or asteroids) from limited observational data. Previous benchmarking of several state-of-the-art 3D reconstruction models on custom spacecraft and asteroid datasets established that Hunyuan3D-2.0 \cite{hunyuan3d22025tencent} consistently outperforms alternative models across both 2D perceptual and 3D geometric metrics. Consequently, we adopt Hunyuan3D-2.0 as well as its lighter, faster variant, Hunyuan3D-mini-turbo, to generate the reconstructed object geometries that serve as the 3D representations for our pose estimation pipeline. 

\subsection{Pose Estimation}
To evaluate the navigational performance of the DreamSat-Bench testbed, we utilize NVIDIA’s FoundationPose (FP) \cite{wen2024foundationpose}. While the platform is designed to be model-agnostic and will eventually incorporate the custom DreamSat-Pose \cite{dreamsat-pose} pipeline, which focuses on monocular learned 2D-3D matching, FP currently serves as a robust, state-of-the-art baseline for 6-DoF tracking. FP is a generalist 6-DoF pose estimator and tracker trained on large-scale synthetic data from diverse 3D object repositories with domain randomization, demonstrating strong zero-shot generalization to unseen objects without fine-tuning. The method utilizes a transformer-based architecture to establish an iterative pose refinement process, where a neural network predicts the relative transformation required to align a 3D model with the observed image features \cite{dreamsat-pose}. For each pose, FoundationPose takes as input an RGB image, an object mask, a depth map, a normalized 3D mesh (.obj), and camera intrinsics. In our setup, all required inputs are provided by DreamSat-Bench. The method of retrieving these inputs is described in Section \ref{sec:dataset-recording}.

\section{DreamSat-Bench: Testbed}
In this section we present the DreamSat-Bench testbed, which is comprised of two primary components: a high-fidelity software simulation environment and a corresponding robotic hardware setup. The software component integrates orbital mechanics with a physics-based rendering engine to simulate both absolute Earth orbits and relative proximity operations between two spacecraft. Within this environment, the chaser spacecraft’s trajectory is traced by a virtual robotic arm whose base is maintained at a fixed relative distance from the target. By utilizing scaled orbits within the simulation, the platform enables a seamless transition of these trajectories to physical robotic hardware, allowing any relative orbit generated in software to be replicated and validated in a HITL setting.

\subsection{Software Setup}
The simulation environment is built on top of tooling developed in \cite{posadasnava2025beavrbimanualmultiembodimentaccessible}. In this setup we target spacecraft RPO. DreamSat-Bench integrates several robotics and machine learning frameworks. MuJoCo \cite{todorov2012mujoco} and serves as the underlying physics engine; MuJoCo Menagerie \cite{Zakka_MuJoCo_Menagerie_A_2022} provides calibrated robot assets (a 7-DOF xArm7 manipulator in the present work). EnvRAANironments are exposed through the standard Gymnasium interface \cite{Towers_Gymnasium_A_Standard}, enabling compatibility with a broad range of policy architectures, or in our case, a computer vision pipeline.

\begin{figure}
    \centering
    \includegraphics[width=0.5\linewidth]{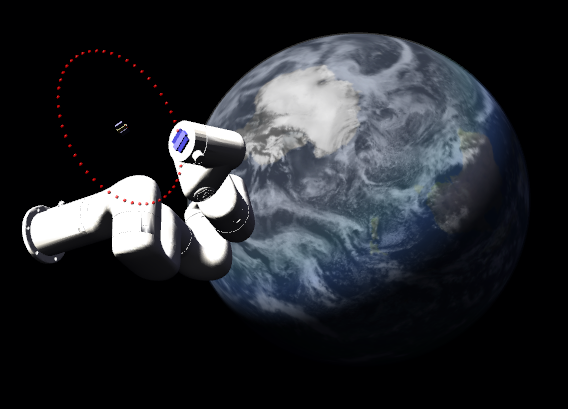}
    \caption{DreamSat-Bench simulation environment. The red dots are the planned robotic motion that trace out a relative orbit around the target spacecraft.}
    \label{fig:sim_env}
\end{figure}

The software is modular. This means we are able to specify the Classical Orbital Elements (COEs) and ROEs used for different mission scenarios. The robotic arm's end-effector represents the chaser spacecraft. Any target body, provided a 3D model is available, can be rendered and fixed in the RTN frame. The simulation decouples orbital dynamics from the physics engine through a two-stage process. In the first stage, orbital mechanics are propagated analytically. For circular orbits, the target spacecraft’s true anomaly advances at the constant mean motion. For elliptical orbits, Kepler’s equation is solved at each timestep to obtain the true anomaly and radius. In the relative-motion case, Gauss’s variational equations cast in ROEs with Hill dynamics generate a pre-computed waypoint sequence in the chief’s local RTN frame. In the second stage, the resulting positions and orientations are kinematically imposed onto the simulation by writing directly to MuJoCo’s joint position array, bypassing actuator dynamics entirely, and propagating the updated generalized coordinates to Cartesian body poses. This architecture cleanly separates trajectory fidelity (governed by the orbital propagator) from rendering and sensing fidelity (governed by the physics engine), allowing the orbital period to be time-scaled arbitrarily without destabilizing the simulator’s integrator. Because the orbital propagator is isolated, a perturbed propagator can be included to model perturbative forces such as $J2$, drag, and solar radiation pressure. The chaser always points its camera toward the target throughout the relative orbit to isolate pose estimation accuracy from relative dynamics.
Earth's rotation and a physically-motivated directional sun light (positioned at 1~AU) produce time-varying illumination conditions, while a high-resolution Earth cubemap texture and a black background provide a realistic visual context \cite{SolarSystemScopeTextures}. Figure \ref{fig:sim_env} shows a rendering of the simulation environment. Note that we render the robotic arm as well as the planned relative orbit, but both can be visually hidden if desired.
The simulation environment consists of a chaser and target spacecraft within a realistic orbital setting (e.g., Earth and celestial backgrounds with varying solar illumination conditions), utilizing a simulated camera with configurable intrinsics (focal length, resolution, and field-of-view). Depth information and segmentation masks of objects are accessible via MuJoCo. Trajectories and object sizes are scaled proportionally to fit within the robotic manipulator's workspace while preserving the geometric relationship between the camera and the target spacecraft. Observation noise is independently configurable: Gaussian noise may be applied to 2-D keypoint detections ($\sigma_\text{px}$) and to the camera orientation ($\sigma_\theta$), allowing evaluation under clean or degraded sensing conditions.
Two independent scale factors govern the scene: a global orbital scale ($s_\text{orb} = 10^{-7}$) maps physical distances to MuJoCo world coordinates, keeping large celestial bodies near the numerical origin to avoid floating-point precision loss at large coordinate magnitudes; and a workspace scale, $s_\text{ws}$, which maps the relative trajectory into the arm's reachable volume:
\begin{equation}
    s_\text{ws} = \frac{\alpha \, d_\text{arm}}{r_\text{orbit}},
\end{equation}
where $d_\text{arm}$ is the XArm7 practical reach and $\alpha$ is a workspace fill fraction calibrated to the simulation geometry. Because the arm is never rescaled, $s_\text{ws}$ must map $r_\text{orbit}$ such that the scaled trajectory falls within the arm’s kinematically feasible region, avoiding singularities near the base and near full extension. The value $\alpha$ was determined empirically to ensure Inverse Kinematics (IK) convergence \cite{Haviland_2024p1, Haviland_2024p2} during the relative motion of the arm around the target. Note that because the chaser’s relative orbit around the target is simulated with a robotic arm, the chaser cannot perform an arbitrary number of orbits around the target in the same direction due to the physical limits of the robot’s joints. To overcome this windup issue, the direction of the relative orbit is alternated after each full revolution (e.g., after the first revolution, the direction of the orbit changes from Counter-Clockwise (CCW) to Clockwise (CW) around the target, then after the second revolution, it changes from CW to CCW, and so on). Even if unphysical, this allows continuous frame extraction to test pose estimation when using hardware. 
Table~\ref{tab:sim_params} summarizes the configurable parameters for the simulation environment. The chief spacecraft orbit is specified by the six COEs in the Earth-Centered Inertial (ECI) frame: semi-major axis $a$, eccentricity $e$, inclination $i$, RAAN $\Omega$, argument of perigee $\omega$, and initial true anomaly $\nu_0$.
Experiments are composed by pairing a camera preset with an orbit preset and a noise level. The camera preset determines the field-of-view (FOV) of the sensor, the focal length ($f$), and the principal points ($c_x, c_y$) reflecting different operational phases of an RPO mission. Table~\ref{tab:cameras} lists the two available camera configurations.
\begin{table}
\small
\centering
\caption{Modular simulation parameters configurable in DreamSat-Bench.}
\label{tab:sim_params}
\begin{tabular}{llll}
\toprule \midrule
Group & Parameter & Default & Description \\
\midrule
\multirow{6}{*}{\shortstack[l]{Chief Orbit\\(ECI / COEs)}}
  & $a$ & $26{,}578\,\text{km}$ & Chief semi-major axis \\
  & $e$ & $0$ & Eccentricity \\
  & $i$ & $51.6\degree$ & Inclination \\
  & $\Omega$ & $0\degree$ & RAAN \\
  & $\omega$ & $0\degree$ & Argument of perigee \\
  & $\nu_0$ & $270\degree$ & Initial true anomaly \\
\midrule
\multirow{4}{*}{ROEs}
& $\delta a$ & 0 & ROE semi-major axis \\
& $\delta\lambda$ & 0 & ROE mean longitude \\
& $\delta e_y$ & 0 & ROE eccentricity \\
& $\delta i_x$ & 0 & ROE inclination \\
& $\delta e_x = \delta i_y$ & $r_\text{orbit}/a$ & In-plane ROE magnitude \\
\midrule
\multirow{1}{*}{Relative Trajectory}
  & $r_\text{orbit}$ & $50$--$1000\,\text{m}$ & Nominal relative orbit size \\
\midrule
\multirow{5}{*}{Scaling}
  & $s_\text{orb}$ & $10^{-7}$ & Global orbital scale (sim-m / real-m) \\
  & $d_\text{arm}$ & $0.7\,\text{m}$ & Arm workspace radius (maximum reach) \\
  & $\alpha$ & $0.11$ & Workspace utilisation factor \\
  & $s_\text{ws}$ & $\alpha\,d_\text{arm}/r_\text{orbit}$ & Workspace scale (derived per preset) \\
\midrule
\multirow{3}{*}{Time}
  & $\tau$ & $500\times$ & Simulation time scale (real-time multiplier) \\
  & Earth rotation & off & Toggle kinematic Earth spin \\
  & Orbital motion & off & Toggle Keplerian orbit propagation \\
\midrule
\multirow{2}{*}{Sensor Noise}
& $\sigma_\text{px}$ & \qty{0}{px} (std.\ dev.) & 2D keypoint translation noise \\
& $\sigma_\theta$ & \qty{0}{\degree} (std.\ dev.) & Camera orientation noise \\
\midrule \bottomrule
\end{tabular}
\end{table}

\begin{table}
\centering
\caption{Simulated camera presets. Both share the same sensor resolution and principal point.}
\label{tab:cameras}
\begin{tabular}{lllllll}
\toprule \midrule
Preset & $f$ [\unit{px}] & Resolution & FOV$_y$ & $(c_x, c_y)$ & Intended range \\
\midrule
Narrow field-of-view (NFOV) & \num{7315} & $1280\times1024$ & $\approx \qty{8.0}{\degree}$  & $(640, 512)$ & \qtyrange{100}{1000}{m} \\
Wide field-of-view (WFOV)   & \num{1545} & $1280\times1024$ & $\approx \qty{36.5}{\degree}$ & $(640, 512)$ & $< \qty{100}{m} $ \\
\midrule \bottomrule
\end{tabular}
\end{table}

\subsection{Perception Pipeline}
\label{sec:perception_pipeline}

The perception pipeline transforms observation sequences produced by the simulation environment into per-frame 6-DoF pose estimates of the target spacecraft. It consists of three sequential stages: dataset recording, single-image 3D reconstruction, and pose tracking. Each stage is implemented such that any 3D reconstruction or pose tracking model can be substituted without modifying the experiment or evaluation code. In this investigation, Hunyuan3D is used for 3D reconstruction. Hunyuan3D requires only images as input, and capably reconstructs from single views. FP is used for pose tracking and offers zero-shot tracking on novel objects and includes model-based and model-free modes. Together, the system results in a training-free, deployment-ready pipeline with operational generalizability.

\subsubsection{Dataset Recording}
\label{sec:dataset-recording}
During simulation, the program writes observations to disk in the LeRobot dataset format \cite{cadene2026lerobot}. Each episode stores the following data at a configurable frame rate:
\begin{enumerate}[topsep=0pt,noitemsep]
\item RGB video frames encoded as videos.
\item A binary segmentation mask video for the target spacecraft.
\item Rendered depth maps stored as per-frame compressed NPZ arrays at full sensor resolution.
\item GT 6-DoF state: target position and quaternion in the world frame, and the $4 \times 4$ camera-to-world transform.
\item Camera intrinsics saved as a matrix.
\item A metadata file encoding the orbital parameters, camera preset, target physical size, and the scaling factor used to convert simulated-meter coordinates back to real-world meters at evaluation time.
\end{enumerate}
Recording is handled asynchronously: rendered frames are placed onto a queue and written to disk by a background thread, so saving data never pauses the simulation. Depth images, which are substantially larger than RGB frames (up to $\sim$1.3 GB per camera at $1280 \times 1024$ resolution), are stored in a memory-mapped file. This is a file on disk that the operating system exposes as if it were RAM, loading only the portions actively being written rather than holding the entire array in memory at once.

\subsubsection{3D Reconstruction}
Given a recorded dataset, a single RGB frame from the stored video is extracted (defaulting to the middle frame if none is specified) and passed to a reconstruction model. The input frame undergoes automatic background removal, as implemented in Hunyuan3D, before being submitted to the model. The reconstructed mesh is then post-processed in two steps:

\begin{enumerate}[topsep=0pt,noitemsep]
\item Scale normalization: Hunyuan3D outputs in its own internal normalized coordinate space (vertices roughly within $[-1, 1]$), so  the mesh is centered at its bounding-box midpoint and uniformly scaled so that its maximum absolute vertex extent matches that of the GT Computer-Aided Design (CAD) reference mesh in real meters. This preserves the physical aspect ratio of the spacecraft.
\item Mesh decimation: If the face count exceeds a target threshold, the mesh is decimated using Open3D quadric decimation \cite{zhou2018open3dmodernlibrary3d}. Trimesh’s built-in decimation is intentionally avoided here, as it silently corrupts non-watertight meshes by collapsing vertices while barely reducing face count, which causes the downstream pose estimator to return degenerate poses. Decimation reduces the memory required to store and transmit the mesh.
\end{enumerate}

The reconstruction stage supports single-view input as well as multi-view input, in which case frames are extracted at configurable indices, and the model is automatically switched to its multi-view variant. Only single-view reconstruction is considered in this work and only a single reconstructed mesh is used per dataset. The pipeline is backend-agnostic.

\subsubsection{Pose Tracking}
Pose estimation is performed frame-by-frame, which coordinates the following steps for each frame in the sequence:

\begin{enumerate}[topsep=0pt,noitemsep]
\item Frame extraction: The RGB frame is extracted from the stored video. Depth is loaded from the pre-indexed NPZ arrays; the entire compressed array is decompressed into RAM on first access, and subsequent frames are served from the in-memory buffer.
\item Coordinate frame conversion: GT poses stored in the dataset use MuJoCo's OpenGL camera convention (X right, Y up, Z backward). The pose estimator, FP, expects the OpenCV convention (X right, Y down, Z forward). The conversion is applied by flipping the Y and Z axes ($\mathbf{T}_\text{cv} = \text{diag}(1,-1,-1,1) \cdot \mathbf{T}_\text{gl}$).
\item Mesh scale computation: The mesh is scaled to physical real-world meters using the target's known physical size from the metadata:
\begin{equation}
    s_\text{real} = \frac{d_\text{target}/2}{\max|\mathbf{v}|}
\end{equation}
where $d_\text{target}$ is the target's physical diameter in meters and $\max|\mathbf{v}|$ is the maximum absolute vertex extent of the normalized mesh. The pose estimator, therefore, operates entirely in real-world meters.
\item Global registration: On the first frame, the pose estimator performs a global registration by generating a rotation hypothesis grid, rendering the mesh from each candidate pose, and scoring matches against the observed RGB and depth using the scorer network. The top hypothesis is then refined iteratively by the refiner network.
\item Sequential tracking: Once initialized, subsequent frames use a lightweight tracking update conditioned on the previous frame's pose. If tracking fails (identity matrix returned), the estimator resets and re-triggers global registration on the next frame. Optionally, periodic re-registration can prevent long-term drift.
\item Depth normalization: FP's internal renderer has a hardcoded $z_\text{far} = 100 \ \text{m}$, and its XYZ normalization assumes tabletop-scale scenes. For orbital ranges beyond $\sim$\qty{50}{m}, a normalization operation rescales the depth and mesh values so that the median object depth maps to 1\,m before calling the estimator, then unscales the translation on output. The rescaling is mathematically exact.
\end{enumerate}

Because FP’s nvdiffrast \cite{laine2020modularprimitiveshighperformancedifferentiable} renderer has a hardcoded far-clip plane, it fails when placing targets beyond close range. To operate at orbital distances up to \qty{500}{m}, a similarity rescaling is applied before every FP call. Given the median target depth $\bar{d}$ over the masked region, a normalizer $k = \frac{\bar{d}}{1 \ \text{m}}$ is computed and applied simultaneously to the depth map and the mesh scale factor: $d' = d/k$, $s' = s/k$. This is a uniform scaling of the entire scene about the camera origin. Because angular subtense is scale-invariant, $\theta = \frac{s}{d} = \frac{s/k}{d/k} = \frac{s'}{d'}$, the object occupies the same pixel footprint before and after rescaling (i.e., FP’s render-and-compare cost function sees an identical matching problem). Rotation is unaffected. The translation FP returns is in the normalized frame; multiplying by $k$ recovers the real-world value exactly: $\mathbf{t}_\text{real} = k\times\mathbf{t}'$.
The rescaling is mathematically lossless. It does not improve the information content of the depth signal; pixel resolution, and depth noise are range-dependent regardless, but it is a prerequisite for the far-range conditions in RQ2 to produce any tracking output at all.

\subsubsection{Evaluation Metrics} 
\label{sec:error_eval}
Pose accuracy is measured as tracking error: The deviation of the estimated pose from GT relative to a first-frame calibration offset. On the first frame, the calibration offset $\Delta \mathbf{T} = \hat{\mathbf{T}}_0^{-1} \mathbf{T}_0^\text{gt}$ is computed to account for any systematic misalignment between the reconstructed mesh coordinate frame and the GT pose. This is only necessary because the pose error is evaluated with respect to a GT pose. However, in a real-world scenario, pose tracking depends on the last registered reference frame. For all subsequent frames $k$, the corrected estimate is defined as $\mathbf{T}_k^{\text{corr}} = \hat{\mathbf{T}}_k \Delta\mathbf{T}$, with translation $\mathbf{t}_k^{\text{corr}}$ and rotation $\mathbf{R}_k^{\text{corr}}$ extracted from $\mathbf{T}_k^{\text{corr}}$, and compared against $\mathbf{T}_k^\text{gt}$:
\begin{equation}
e_t^{(k)} = \left| \mathbf{t}_k^{\text{corr}} - \mathbf{t}_k^\text{gt} \right|_2 , \quad
e_r^{(k)} = \left| \log\left( (\mathbf{R}_k^{\text{corr}})^\top \mathbf{R}_k^\text{gt} \right) \right|
\end{equation}
where $e_t$ is expressed in real-world meters and $e_r$ is expressed in degrees.

\subsection{Hardware Setup}
\label{sec:hardware-test-setup}
The hardware suite is comprised of one 7-DoF xArm manipulator, a workstation equipped with AMD Ryzen 9 9900X CPU 4.40 GHz processor, an NVIDIA GeForce RTX 4090, 32.0 GB of RAM, running a 64-bit Windows 11 operating system, an Orbbec 305g Stereo-Vision Camera, a Amaran Ray 120c light source with Amaran Mini Fresnel attachment, and a camera tripod interacting with 3D-printed models of the chaser and target spacecraft.
To prepare the hardware testing environment a frame was constructed (8’ long x 4’ wide x 4’ tall) out of extruded aluminum rails. Black-out curtains were then used to enclose the frame and reduce light entering the enclosure. Within the enclosure, the xArm7 robotic manipulator was installed with the Orbbec 305g Camera on the end-effector. The camera tripod was placed in the enclosure so as to align closely with the simulated position of the target spacecraft. This is where the physical, 3D printed spacecraft models were placed during testing. Finally, the Amaran Ray 120c lightsource with fresnel lens attachment was mounted within the enclosure so as to illuminate the target in a similar manner as the sun illuminates an on-orbit spacecraft. 
\begin{figure}
    \centering
        \centering
        \includegraphics[width=0.55\textwidth]{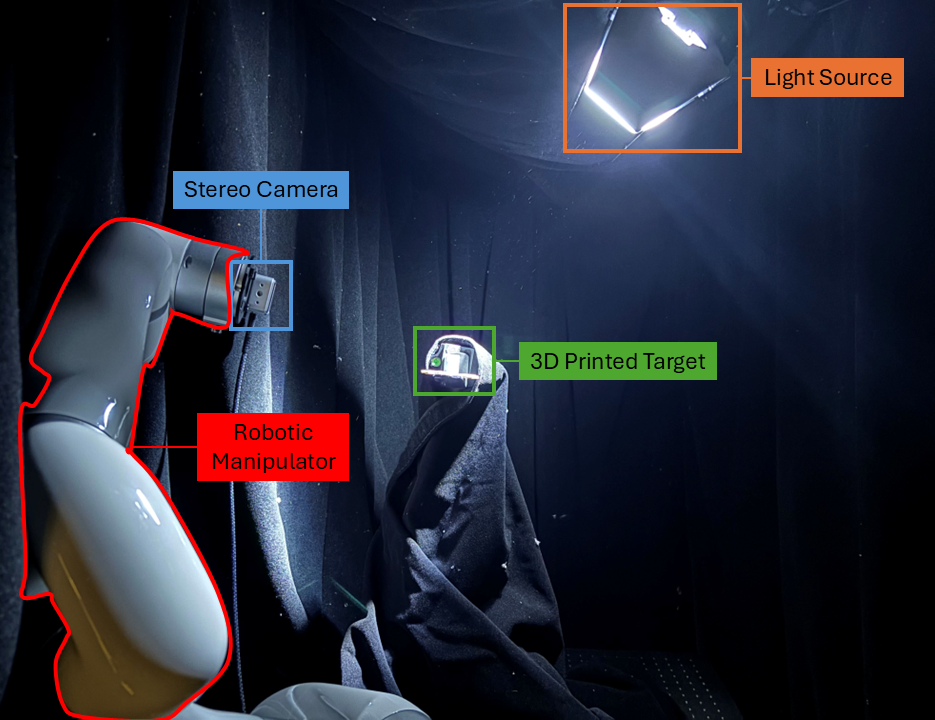}
        \label{fig:hardware-interiror}
    \caption{Hardware test enclosure.}
    \label{fig:hardware-interior}
\end{figure}

\section{Experimental Campaigns}
This section details the experimental methodology used to benchmark the proposed navigation framework, encompassing a multi-parameter SITL study and a preliminary HITL validation.

\subsection{Experiment Test Campaign}

Five spacecraft models are evaluated using datasets generated in the MuJoCo simulation environment. These models were taken from the Dreamsat-2.0 \cite{dreamsat2} validation dataset used in a previous work . Table \ref{fig:reconstruction-comparison} shows the selected spacecraft models, chosen for varying geometries and degree of symmetry. Specifically, models 3, 4, and 5 have two planes of symmetry, model 2 has 1 plane of symmetry and model 1 exhibits no symmetry. Each dataset contains synchronized RGB images, depth maps, segmentation masks, and simulated GT target poses. FP is used to estimate the target pose from either a reconstructed mesh or the corresponding GT mesh. Performance is quantified using translational and rotational pose error relative to the simulated GT. Unless otherwise stated, each experimental condition is evaluated over a 150-frame sequence, and the GT mesh is evaluated as a baseline.

\subsubsection{Research Question 1 (RQ1): Reconstruction Time}
How does pose-estimation accuracy vary with reconstruction time, and at what point do additional reconstruction steps produce diminishing returns?
For each spacecraft model, a one-orbit dataset is generated with a nominal 50 m separation between the chaser and target. The trajectory is defined by the COEs and ROEs described in Table \ref{tab:sim_params}. ROEs have been chosen such that there is no drift in the relative orbit. It should be noted that the relative orbit itself is elliptical, and thus the 50 m range is non-constant throughout the orbital period. A single image corresponding to an oriented Sun phase angle of 68.0° is selected from each dataset for reconstruction, ensuring consistent illumination across the spacecraft models. Throughout this work, a phase angle of $0\degree$ represents the chaser located directly between the Sun and target while a phase angle of $180\degree$ represents the target located directly between the Sun and chaser. For the purposes of plotting data, a full $360\degree$ range is used rather than the conventional $0\degree-180\degree$. Values $0\degree-180\degree$ signify the region where the spacecraft approaches occultation while values $180\degree-360\degree$ signify the region where the spacecraft leaves occultation.

Reconstructions are generated using both the Hunyuan3D-2.0 full model and the Hunyuan3D mini-turbo model. Six inference-step settings are evaluated for each model architecture. During each inference step, the diffusion transformer predicts a velocity in latent shape space, and an Euler scheduler uses that prediction to update the latent representation toward an image-conditioned 3D shape. After the final step, the latent representation is decoded and its surface extracted as a mesh. Increasing the number of inference steps provides a finer approximation of this generative trajectory while increasing the computation time.
The measured reconstruction time is recorded for every generated mesh. FP then evaluates each reconstruction over the same 150-frame interval, corresponding to an oriented Sun phase angle change from approximately 68.0° to 23.2°. The resulting pose estimates are compared with the simulated target poses to calculate translational and rotational error. The GT mesh is also used for pose estimation with FP to establish a baseline independent of reconstruction error.

\subsubsection{Research Question 2 (RQ2): Mesh Resolution}
How does reconstructed-mesh resolution affect pose-estimation accuracy? For each spacecraft model, the best-performing reconstruction identified in RQ1 is selected for evaluation. Each selected reconstruction is decimated to target face counts of 50,000, 10,000, 5,000, 1,000, and 100 faces. This approach varies mesh complexity while holding the underlying reconstructed geometry and evaluation sequence constant.
Each decimated mesh is supplied to FP and evaluated using the corresponding 50 m dataset and 150-frame sequence from RQ1. The original GT mesh is evaluated under the same conditions as a baseline. Translational and rotational pose errors are calculated relative to the simulated target pose and aggregated for each target face count, allowing the effect of mesh resolution on pose-estimation accuracy to be evaluated.

\subsubsection{Research Question 3 (RQ3): Orbital Range}
How does chaser–target range affect the accuracy of the reconstruction-based pose-estimation pipeline?
For each spacecraft model, one-orbit datasets are generated at chaser–target separations of 5, 10, 25, 50, 100, 150, 200, and 250 m. The target size is held constant at 2 m. The orbital parameters associated with each range are provided in Table \ref{tab:sim_params}. A wide-field-of-view camera configuration is used at ranges of 5–50 m, while a narrow-field-of-view configuration is used at ranges of 100–250 m. The corresponding camera intrinsics are provided in Table \ref{tab:cameras}. The same best-performing reconstruction selected from RQ1 is used at every range for a given spacecraft, thereby isolating the effect of observation distance and camera configuration.
FP is used to evaluate pose estimates over the same 150-frame orbital interval at each nominal range. Both the reconstructed and GT meshes are tested, allowing reconstruction-induced error to be distinguished from the range-dependent behavior of the pose-estimation system. Translational and rotational pose errors are calculated relative to the simulated target pose and aggregated for each orbital condition.

\subsubsection{Research Question 4 (RQ4): Illumination Condition}
How does pose-estimation accuracy vary with on-orbit illumination geometry?
The 50 m one-orbit datasets are divided into five non-overlapping 150-frame evaluation intervals spanning different illumination conditions. Sun phase angle is represented as an oriented angle from 0° to 360°, where 0° places the chaser on the target-to-Sun direction, with the chaser between the Sun and target, and 180° places the target between the Sun and chaser.
For each spacecraft model, FP is used to evaluate both the best-performing reconstruction identified in RQ1 and the corresponding GT mesh. Each phase-angle interval is treated as an independent tracking run so that initialization and tracking behavior can be evaluated consistently across illumination conditions. Pose estimates are compared with the simulated target poses to calculate translational and rotational errors as functions of Sun phase angle.

\subsection{Hardware}
The hardware test campaign is primarily a preliminary test to establish the environment, configuration, and control software for further tests using physical hardware rather than a simulated environment. The following section describes the established test procedure, current limitations, and future plans for the hardware testbed. 

\subsubsection{Hardware Test Procedure}
The hardware test environment is configured as described in Section~\ref{sec:hardware-test-setup}. During testing, first, we use the Mujoco simulation environment to generate a trajectory NPZ file that encodes the joint movements resulting in a path that is physically representative of orbital dynamics and still achievable by the constraints of the xArm7’s range and obstacles imposed by the enclosure and tripod mount. The trajectories currently describe a partial ellipse around the target model, with the end-effector camera oriented such that it is always pointing towards the target. 
Once the trajectory file is generated, a python script is run that simultaneously moves the arm through the trajectory and records both RGB images and depth data using the Gemini 305g camera. Optionally, the Mujoco simulation can run concurrently to gather GT information of the simulated model, for later comparison to the physical captured data. After the data collection phase is complete, one or more frames are selected to use for the image-to-3D model reconstruction. The primary reconstruction model used is the Hunyuan3D-2.0 as well as the mini-turbo variant. 
The reconstructed model as well as the RGB image and depth information are given to FP for the pose estimation phase. Optionally, the GT 3D model can be used for the pose estimation step, to get a baseline for performance using real image/depth inputs. After pose estimation, that data is compiled and the hardware test pipeline is complete. 

\subsubsection{Limitations and Future Work}
The first limitation to the current pipeline is the lack of a GT pose with which to compare the estimates and determine pose error. Currently, the pose estimates are evaluated qualitatively to see how well the pose estimate tracks visually. This is because the pose of the physical target object is not yet accurately calibrated to match the simulation environment. There are two approaches to how this calibration can be accomplished in future work. 
The first is to use a second robotic manipulator that can be controlled precisely using the same Mujoco environment. If the target is accurately mounted to the second robotic manipulator, then the target’s GT pose can be retrieved from the simulation environment and used to calculate pose error. The alternative strategy is to use high-fidelity optical tracking devices, such as OptiTrack cameras. Markers would be placed onto the target model, and the OptiTrack cameras could provide a high-accuracy pose estimate that is used as an approximate GT pose for comparison.
The second limitation is the spacecraft models used. While 3D printed models can provide a wide variety of geometries quickly and cheaply, they are limited in how accurately they can simulate a real on-orbit spacecraft. In future work, custom or commercial scale-model spacecraft can be used to increase the level of realism in these hardware tests.

\section{Simulation and Hardware Results}

This section evaluates DreamSat-Bench through four SITL experiments and a preliminary hardware demonstration. Across five spacecraft models, the simulation campaign examines how reconstruction time, mesh resolution, chaser--target range, and illumination geometry affect FP tracking, using the corresponding GT meshes as baselines. The hardware campaign qualitatively validates the physical RGB-D acquisition, reconstruction, and pose-tracking pipeline; quantitative errors are not yet reported because the physical target pose has not been calibrated in 6-DoF. Together, these experiments characterize the principal accuracy--computation trade-offs and tracking failure modes of the proposed pipeline.

\subsection{Simulation Experiment: Reconstruction Time}

\begin{figure}[H]
    \centering

    \begin{subfigure}[t]{0.48\textwidth}
        \centering
        \includegraphics[width=\linewidth]{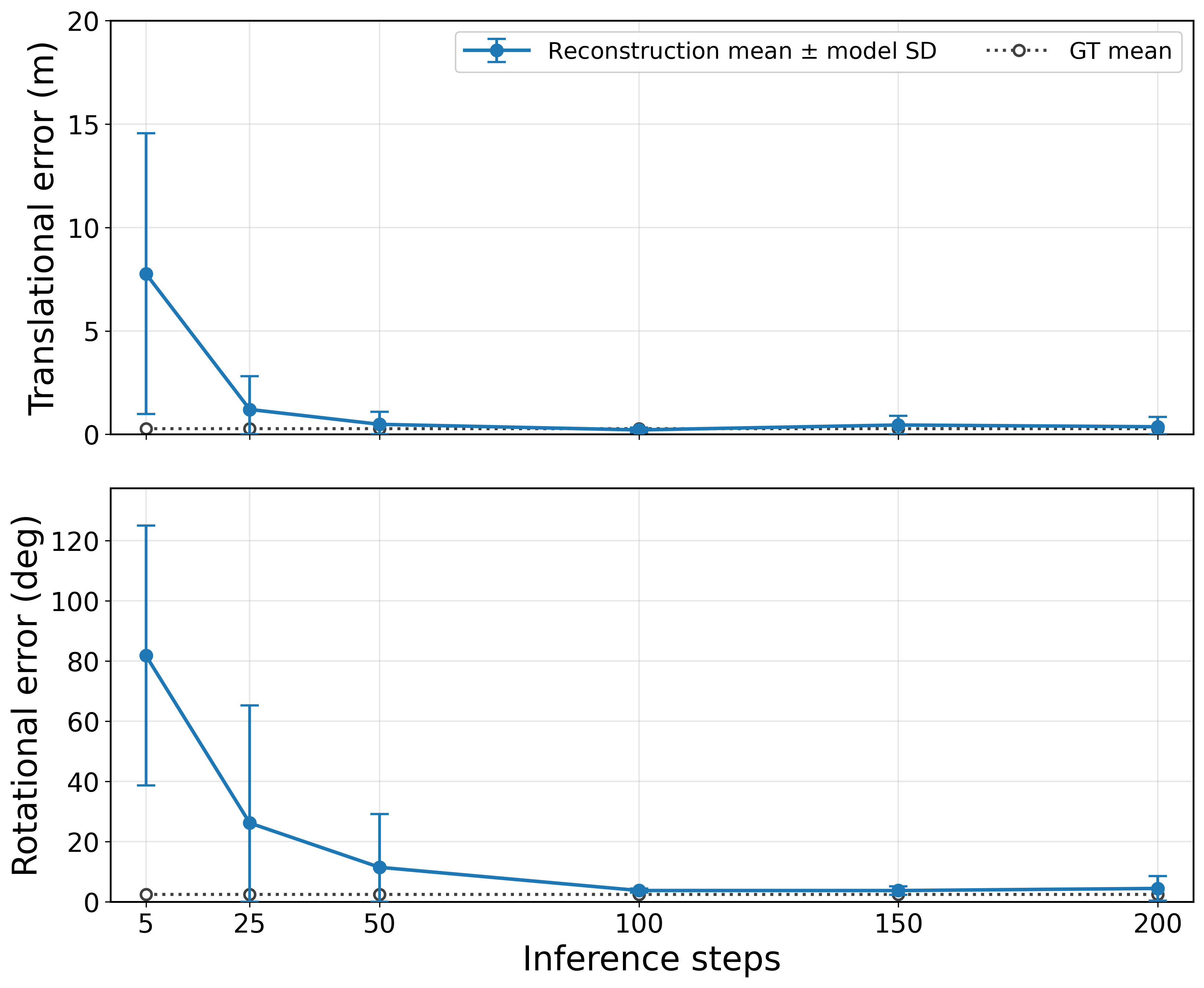}
        \caption{Hunyuan3D-2.0 Full.}
        \label{fig:rq1-full}
    \end{subfigure}
    \hfill
    \begin{subfigure}[t]{0.48\textwidth}
        \centering
        \includegraphics[width=\linewidth]{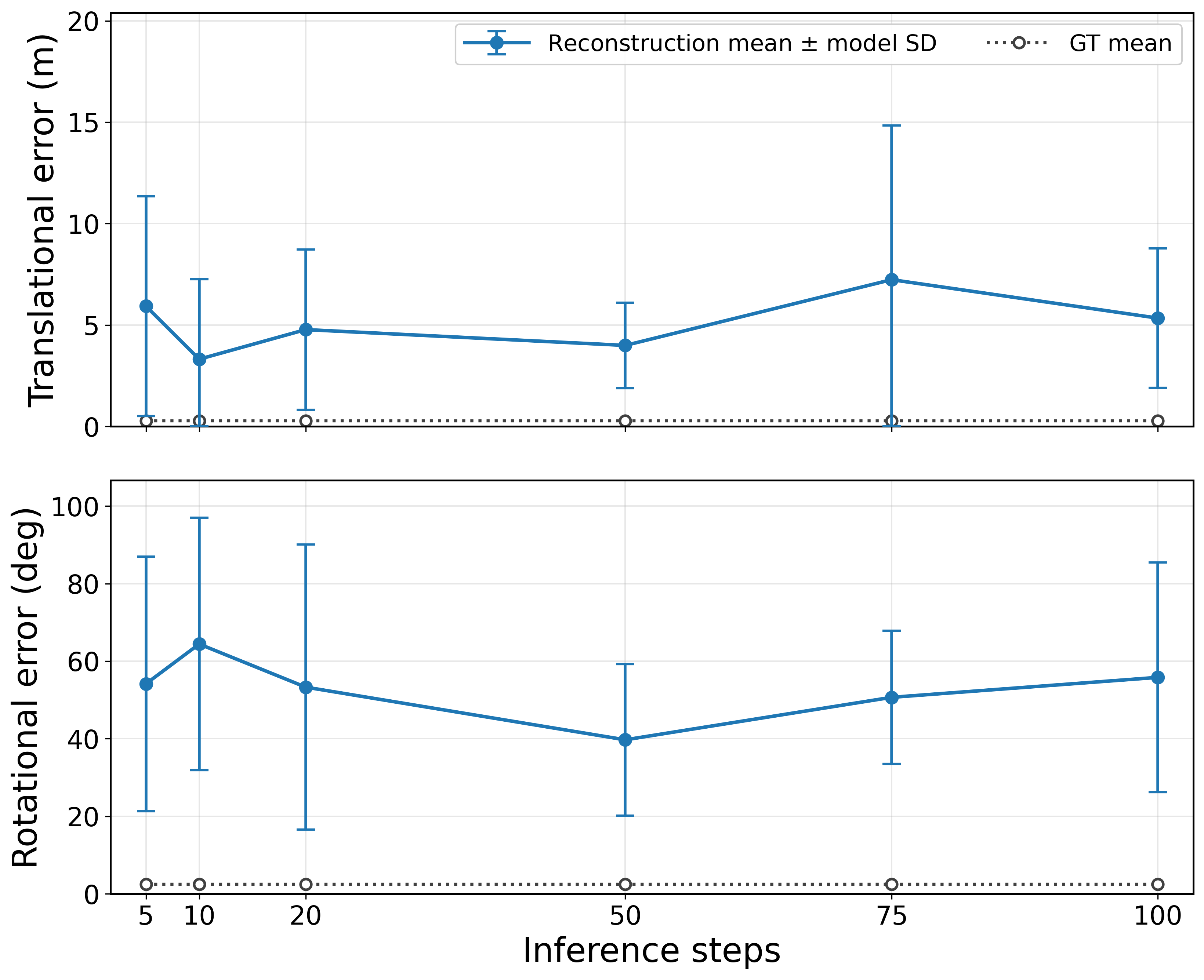}
        \caption{Hunyuan3D-2.0 Mini-Turbo.}
        \label{fig:rq1-mini}
    \end{subfigure}

    \caption{Mean translational and rotational pose-estimation error as a
    function of reconstruction time for the Hunyuan3D-2.0 Full and
    Mini-Turbo reconstruction models.}
    \label{fig:rq1-reconstruction-time-comparison}
\end{figure}
Figure~\ref{fig:rq1-full} shows the measured pose error as a function of reconstruction time for each of the five spacecraft models for the Hunyuan3D Full reconstruction model. At the 100-step setting, which requires approximately 17.5--17.9~s per reconstruction on the computing hardware described in \ref{sec:hardware-test-setup}, all five models achieve mean translational errors below 0.341~m and mean rotational errors below $4.63\degree$. Relative to the 5-step condition, the average translational and rotational errors across the five models decrease by 97.3\% and 95.4\%, respectively. At the 50-step setting, corresponding to reconstruction times of approximately 9.3--9.7~s, four of the five models achieve errors below 0.30~m and $5\degree$. All five models simultaneously achieve errors below 0.5~m and $5\degree$ only at the 100-step setting. Increasing the number of inference steps beyond this point does not consistently improve pose-estimation accuracy, indicating diminishing and non-monotonic returns from additional reconstruction time.
The Hunyuan3D-Mini-Turbo model is evaluated using the same reconstruction and pose-estimation pipeline and results are shown in Figure \ref{fig:rq1-mini}. The Mini-Turbo model enforeces a maximum inference step count of 100 steps. At the 100-step maximum, Mini-Turbo requires approximately 7.1--7.3~s per reconstruction, compared with 17.5--17.9~s for the full model. This represents an approximately 59\% reduction in reconstruction time, or a speedup of approximately 2.4$\times$. 
This computational advantage is accompanied by substantially higher downstream pose-estimation error. At 100 steps, Mini-Turbo produces average translational and rotational errors of 5.34~m and $55.8\degree$, respectively, compared with 0.212~m and $3.75\degree$ for the full model. The Mini-Turbo errors are therefore approximately 25 times larger in translation and 15 times larger in rotation. Furthermore, its lowest cross-model mean translational error is 3.32~m at 10 steps, whereas its lowest cross-model mean rotational error was $39.7\degree$ at 50 steps. The occurrence of these minima at different step counts demonstrates that increasing the number of inference steps does not consistently improve both error metrics. None of the 30 tested Mini-Turbo model--step combinations simultaneously achieved mean errors below 0.5~m and $5\degree$. These results indicate that, for the spacecraft and viewing conditions evaluated, Mini-Turbo's reduction in reconstruction time came at the cost of substantially less reliable downstream pose estimation.
Figure \ref{fig:reconstruction-comparison} below shows a comparison of the input images taken from the MuJoCo simulation environment, the reconstructed 3D models itself, and the GT 3D models used for this test.

\begin{figure}[H]
    \centering
    \small

    \begin{tabular}{@{} r c c c c @{}}

        \textbf{Model 1} &
        \includegraphics[width=0.19\textwidth,height=1.8cm,keepaspectratio]
        {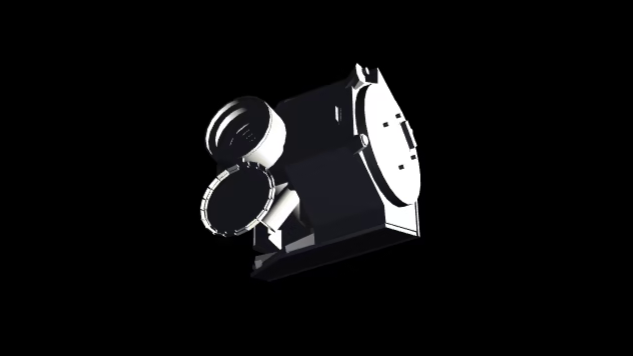} &
        \includegraphics[width=0.19\textwidth,height=1.8cm,keepaspectratio]
        {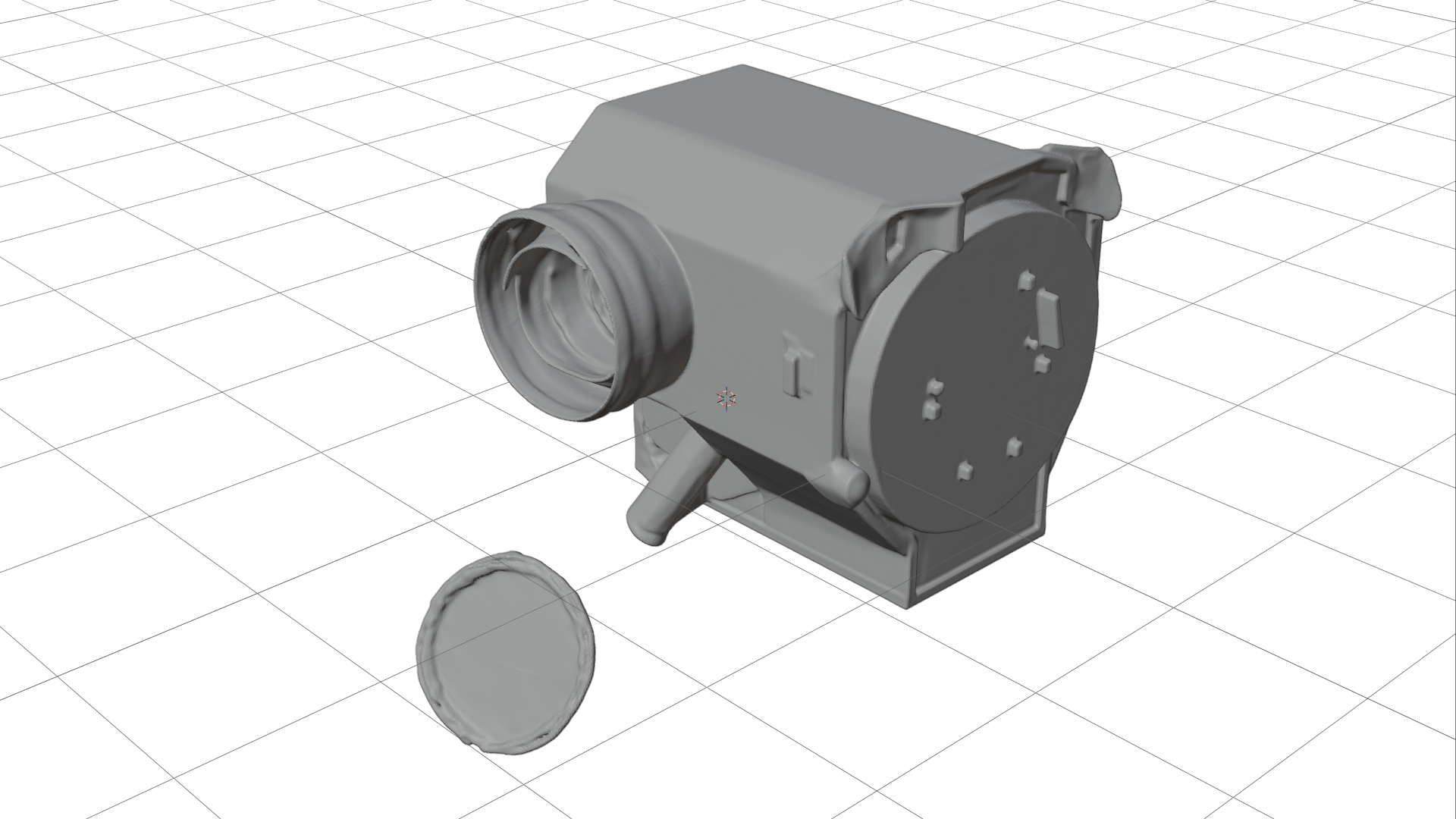} &
        \includegraphics[width=0.19\textwidth,height=1.8cm,keepaspectratio]
        {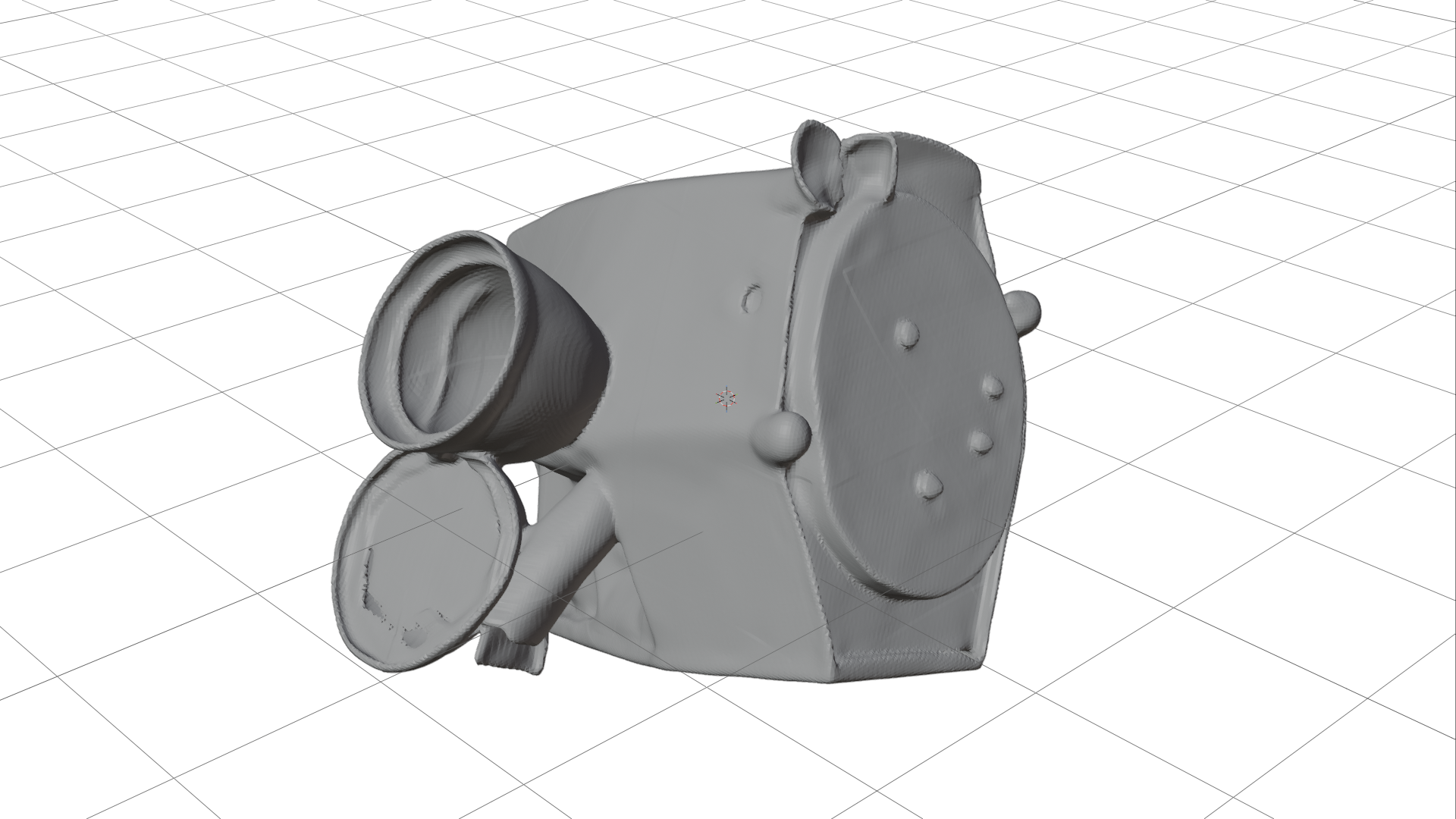} &
        \includegraphics[width=0.19\textwidth,height=1.8cm,keepaspectratio]
        {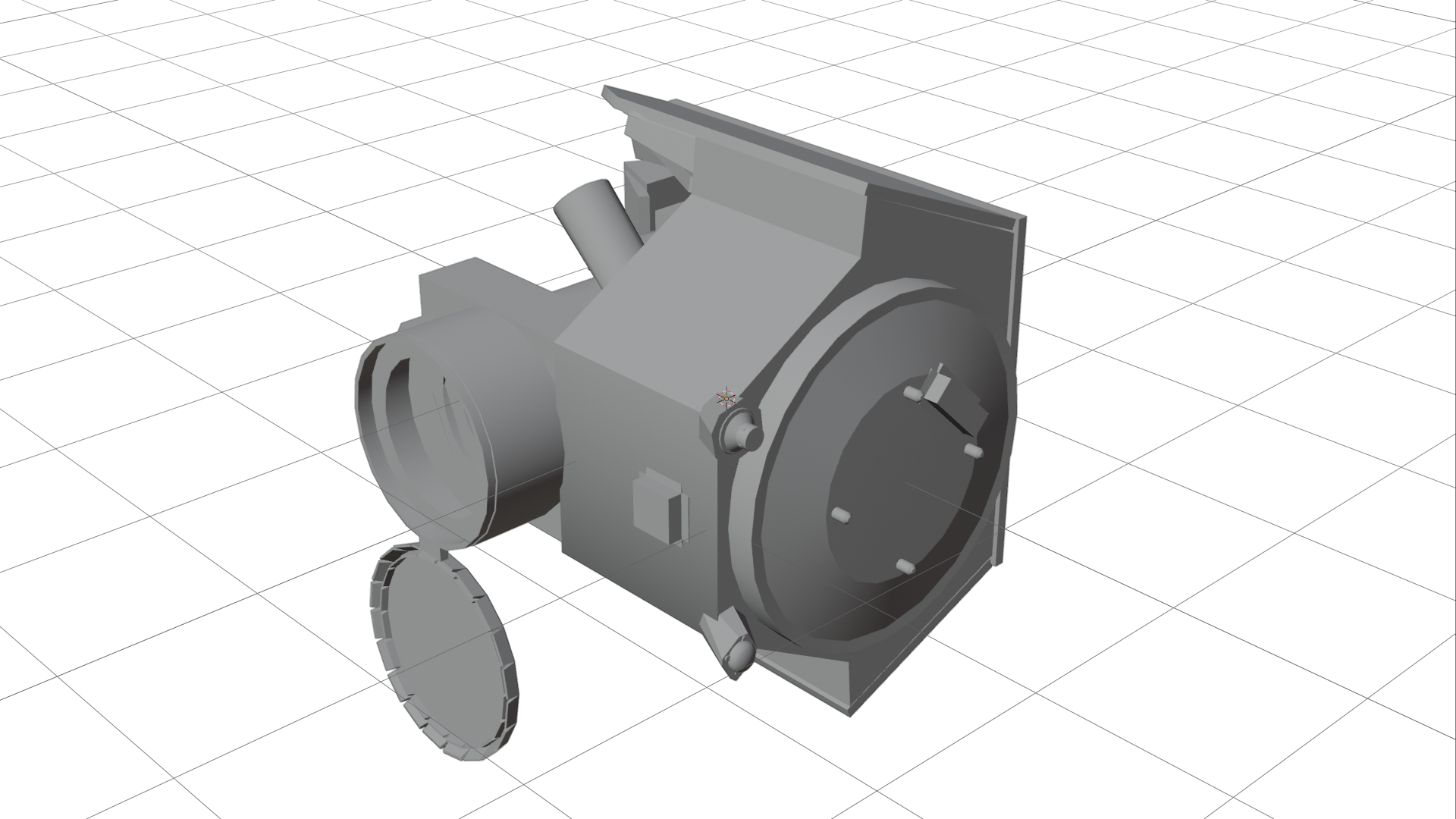}
        \\[1mm]

        \textbf{Model 2} &
        \includegraphics[width=0.19\textwidth,height=1.8cm,keepaspectratio]
        {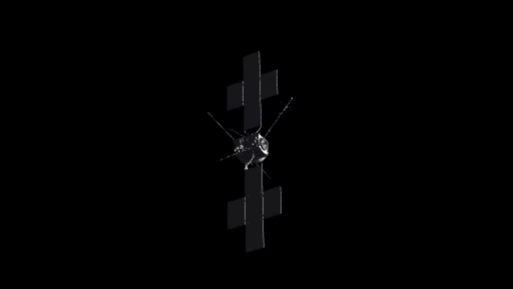} &
        \includegraphics[width=0.19\textwidth,height=1.8cm,keepaspectratio]
        {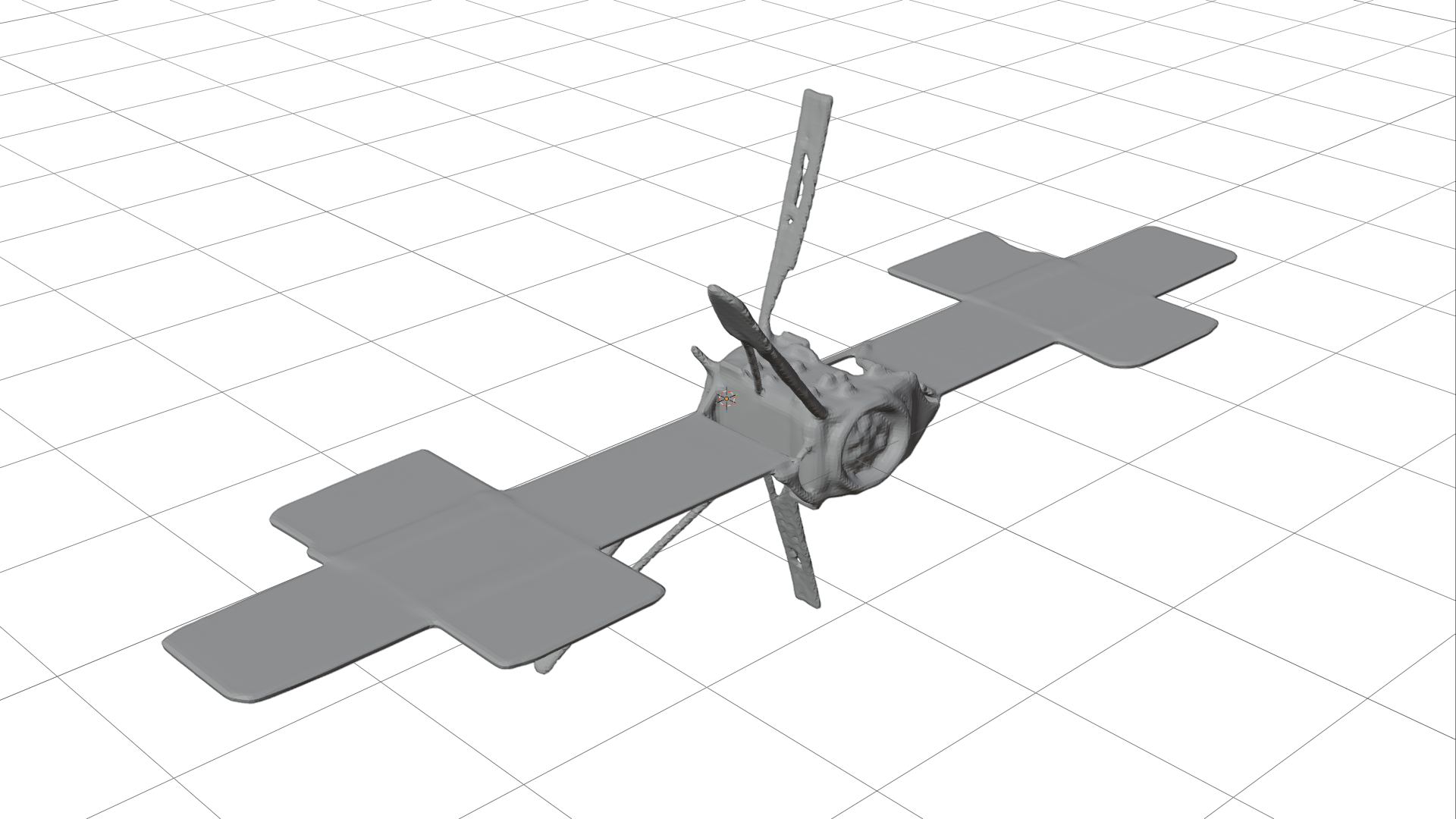} &
        \includegraphics[width=0.19\textwidth,height=1.8cm,keepaspectratio]
        {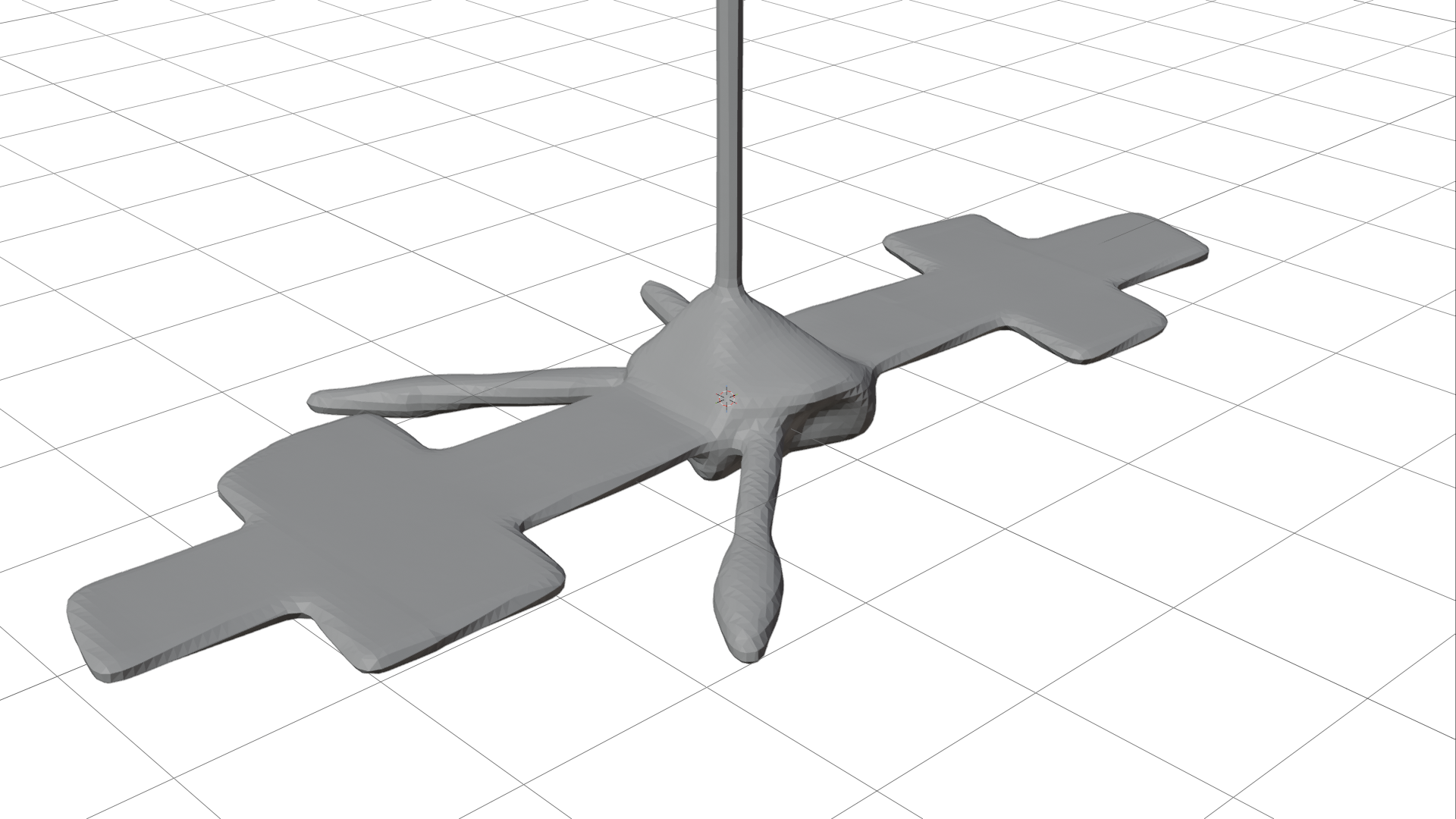} &
        \includegraphics[width=0.19\textwidth,height=1.8cm,keepaspectratio]
        {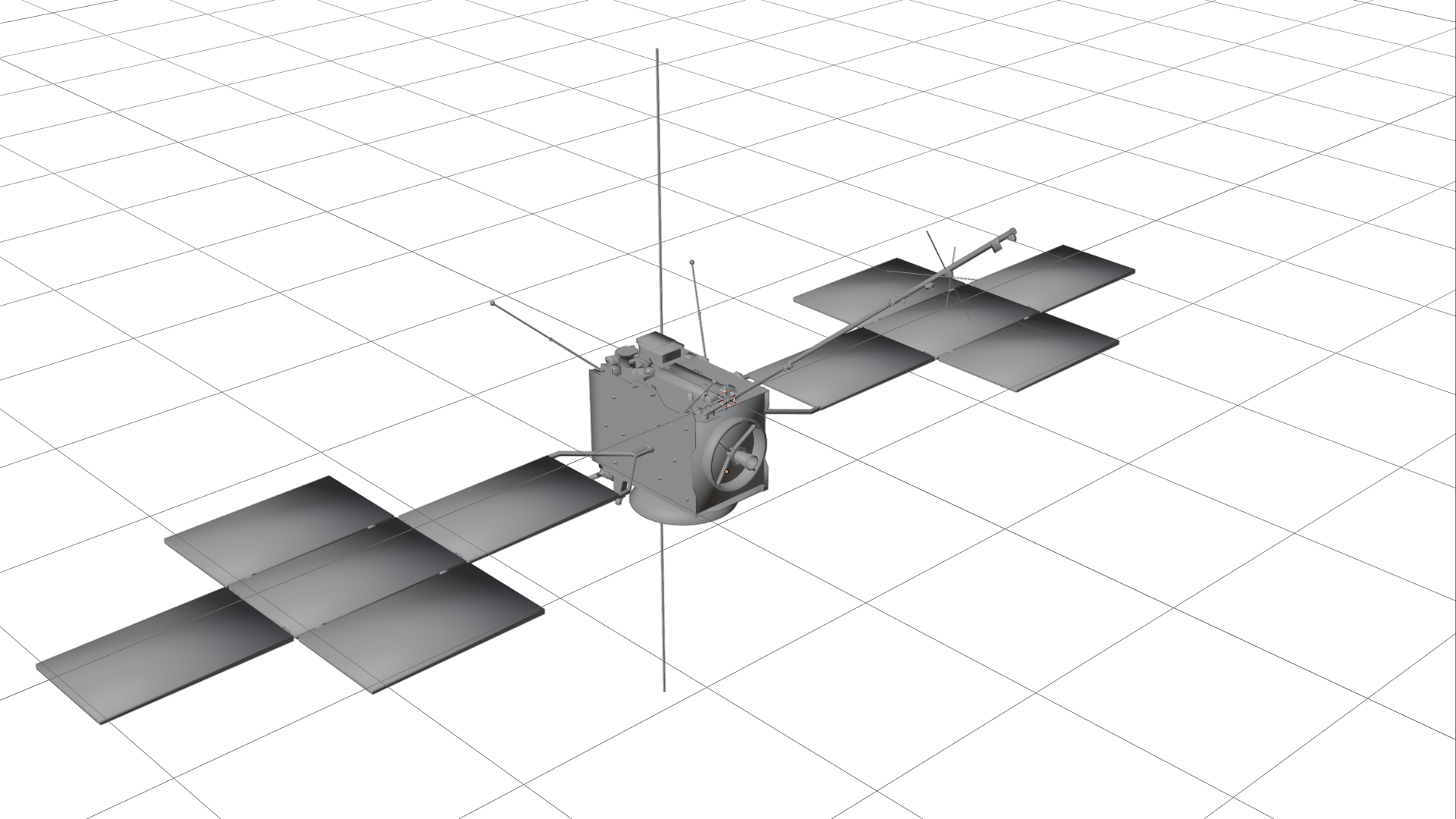}
        \\[1mm]

        \textbf{Model 3} &
        \includegraphics[width=0.19\textwidth,height=1.8cm,keepaspectratio]
        {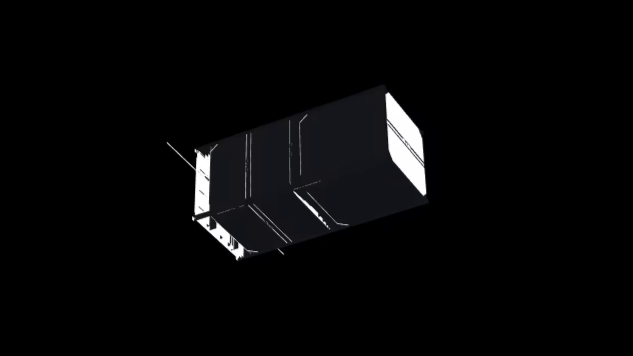} &
        \includegraphics[width=0.19\textwidth,height=1.8cm,keepaspectratio]
        {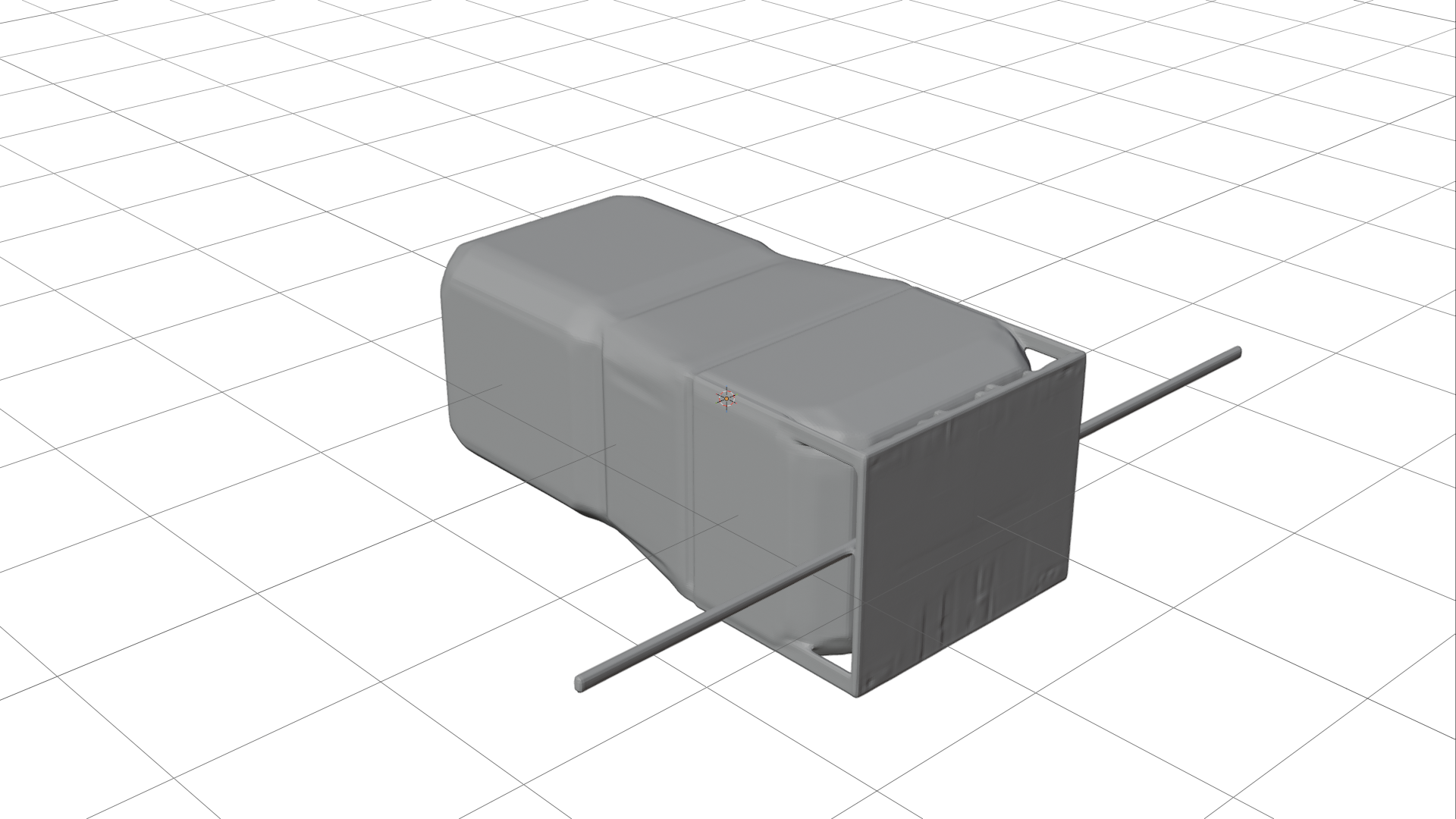} &
        \includegraphics[width=0.19\textwidth,height=1.8cm,keepaspectratio]
        {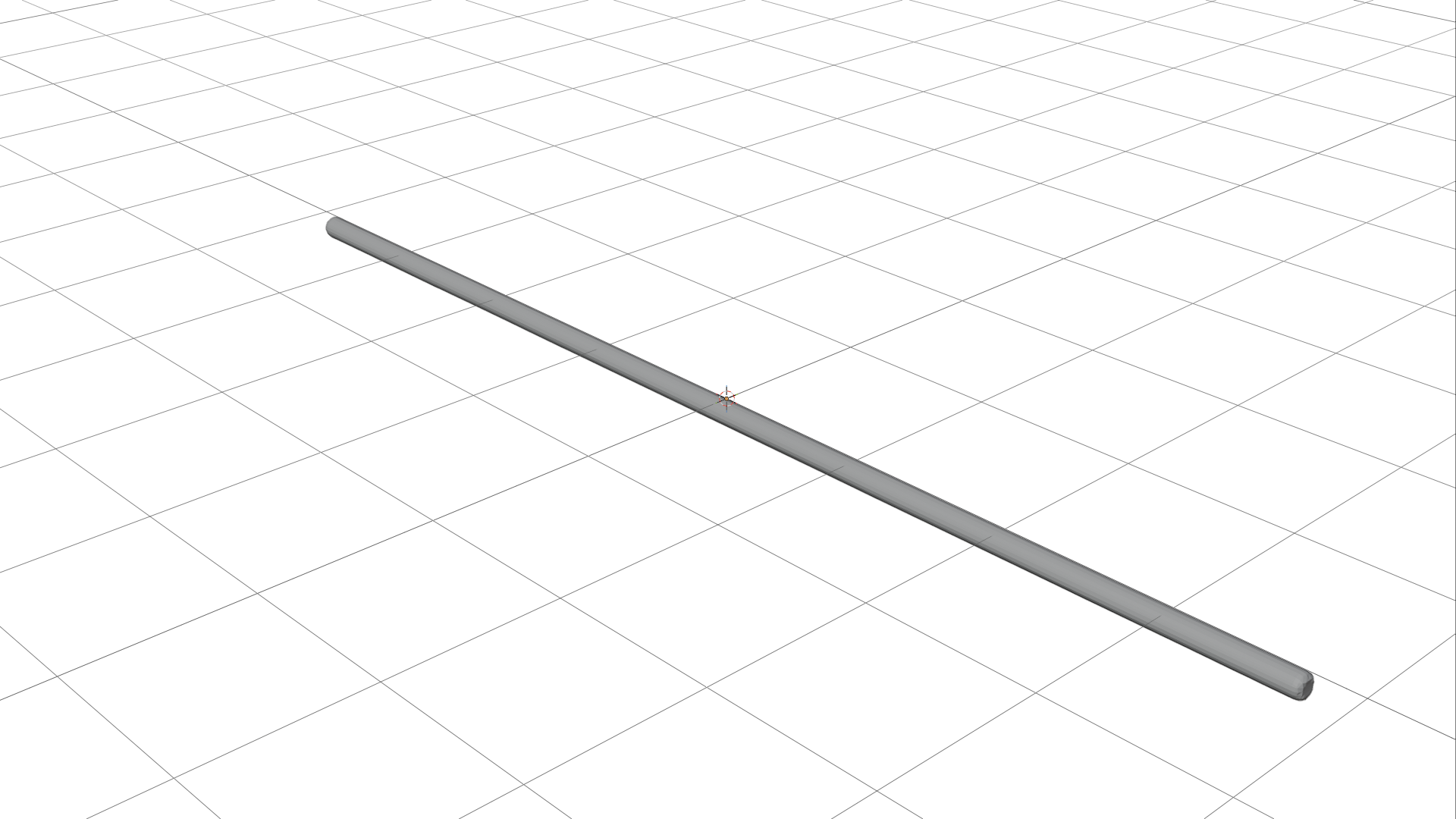} &
        \includegraphics[width=0.19\textwidth,height=1.8cm,keepaspectratio]
        {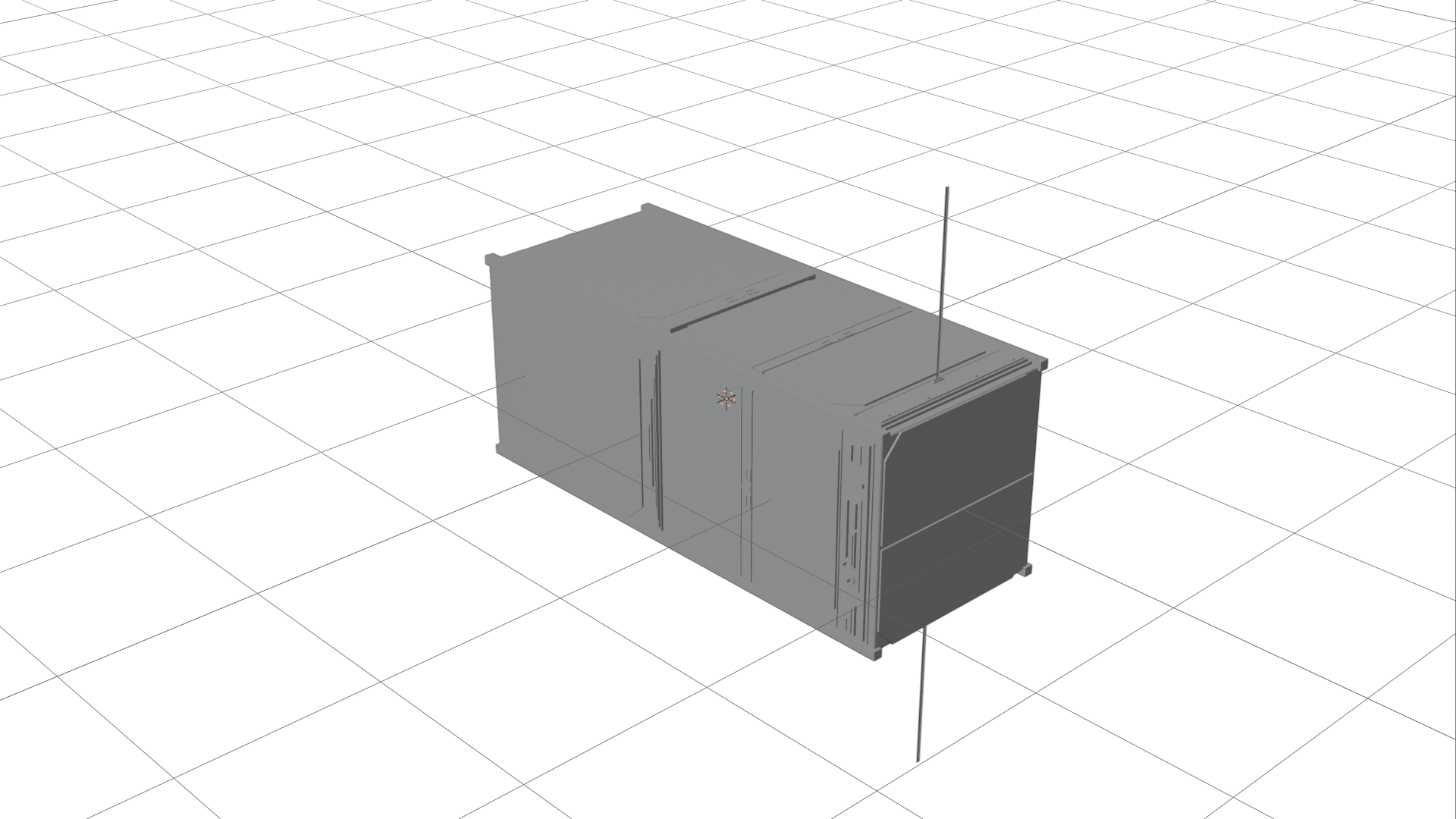}
        \\[1mm]

        \textbf{Model 4} &
        \includegraphics[width=0.19\textwidth,height=1.8cm,keepaspectratio]
        {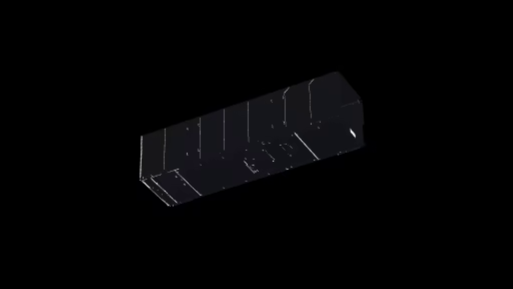} &
        \includegraphics[width=0.19\textwidth,height=1.8cm,keepaspectratio]
        {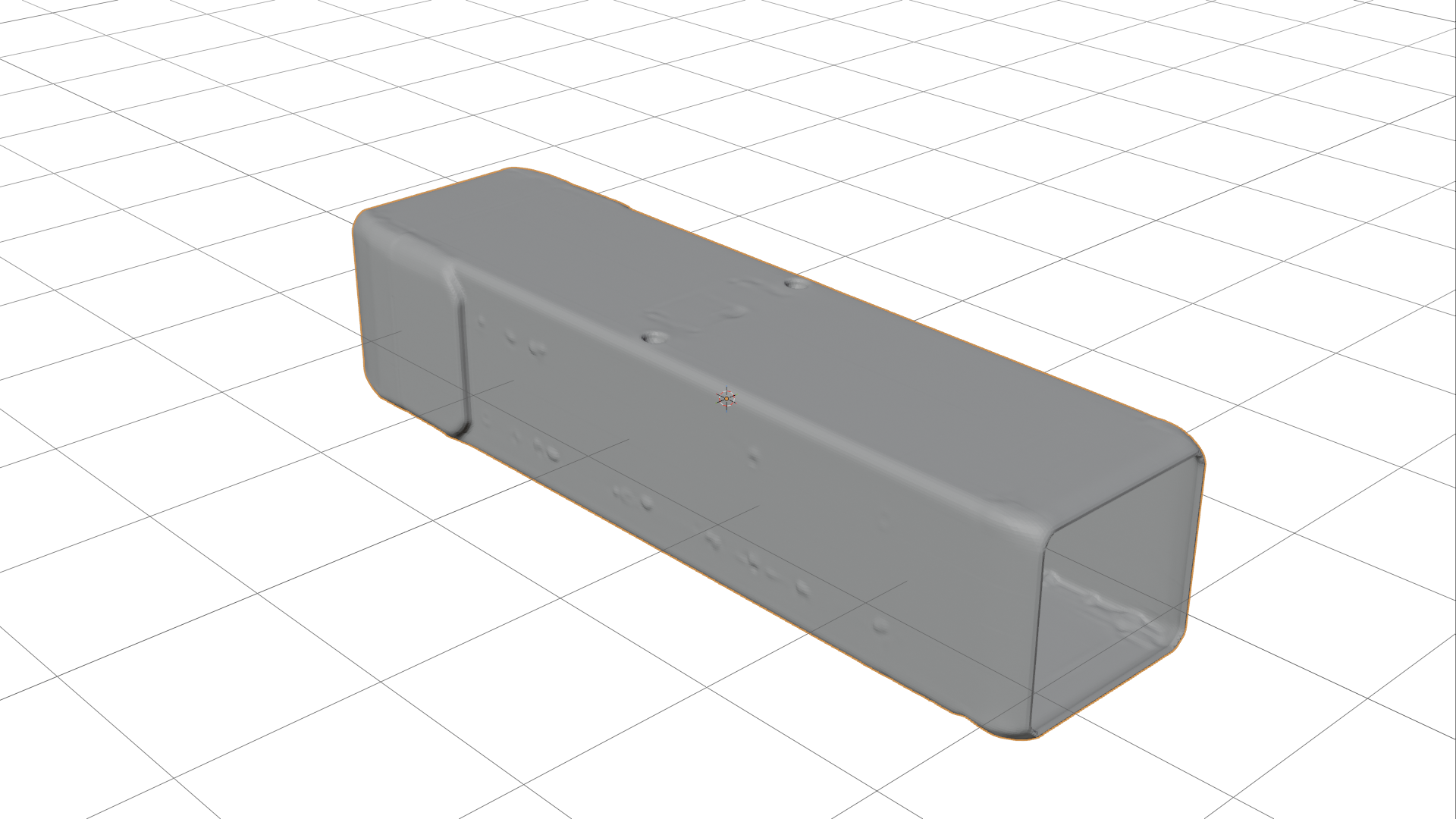} &
        \includegraphics[width=0.19\textwidth,height=1.8cm,keepaspectratio]
        {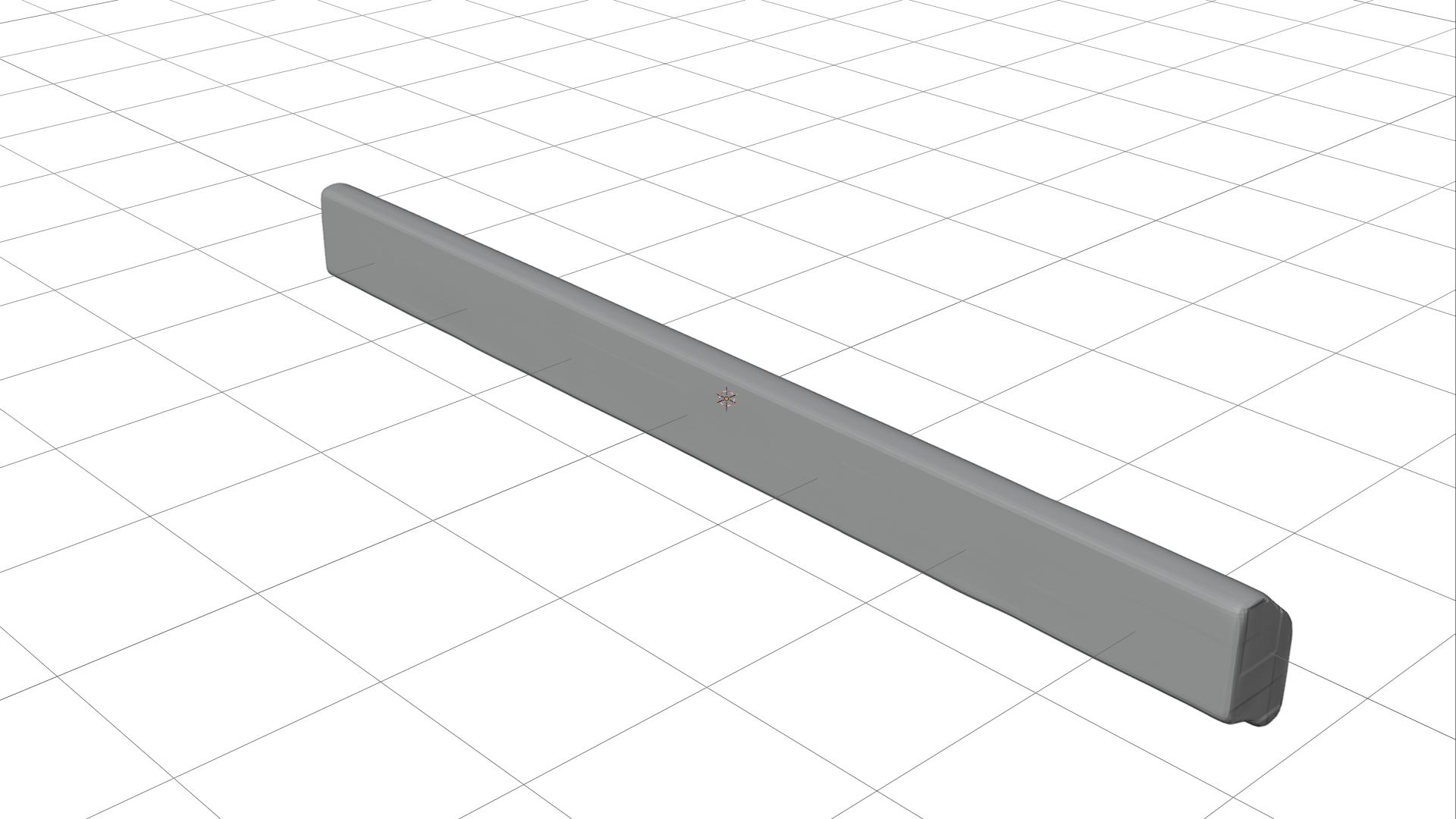} &
        \includegraphics[width=0.19\textwidth,height=1.8cm,keepaspectratio]
        {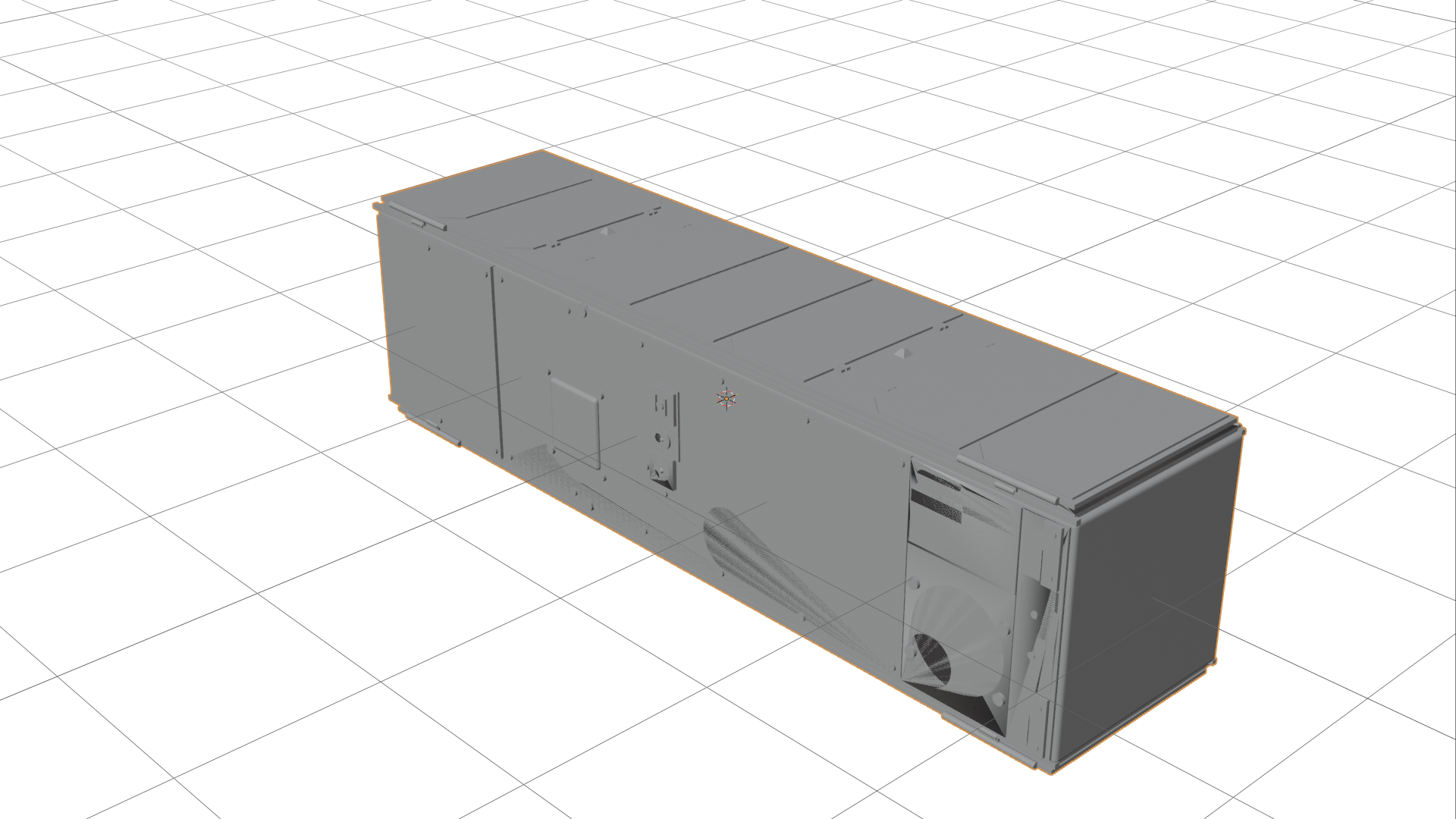}
        \\[1mm]

        \textbf{Model 5} &
        \includegraphics[width=0.19\textwidth,height=1.8cm,keepaspectratio]
        {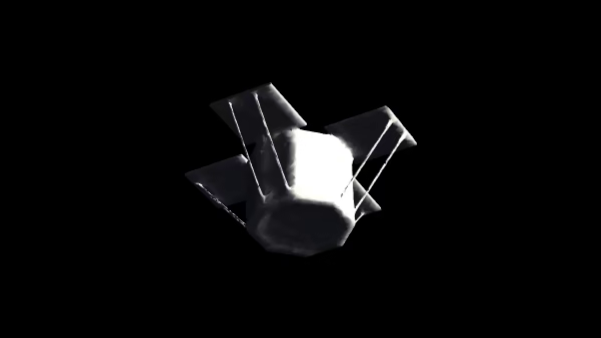} &
        \includegraphics[width=0.19\textwidth,height=1.8cm,keepaspectratio]
        {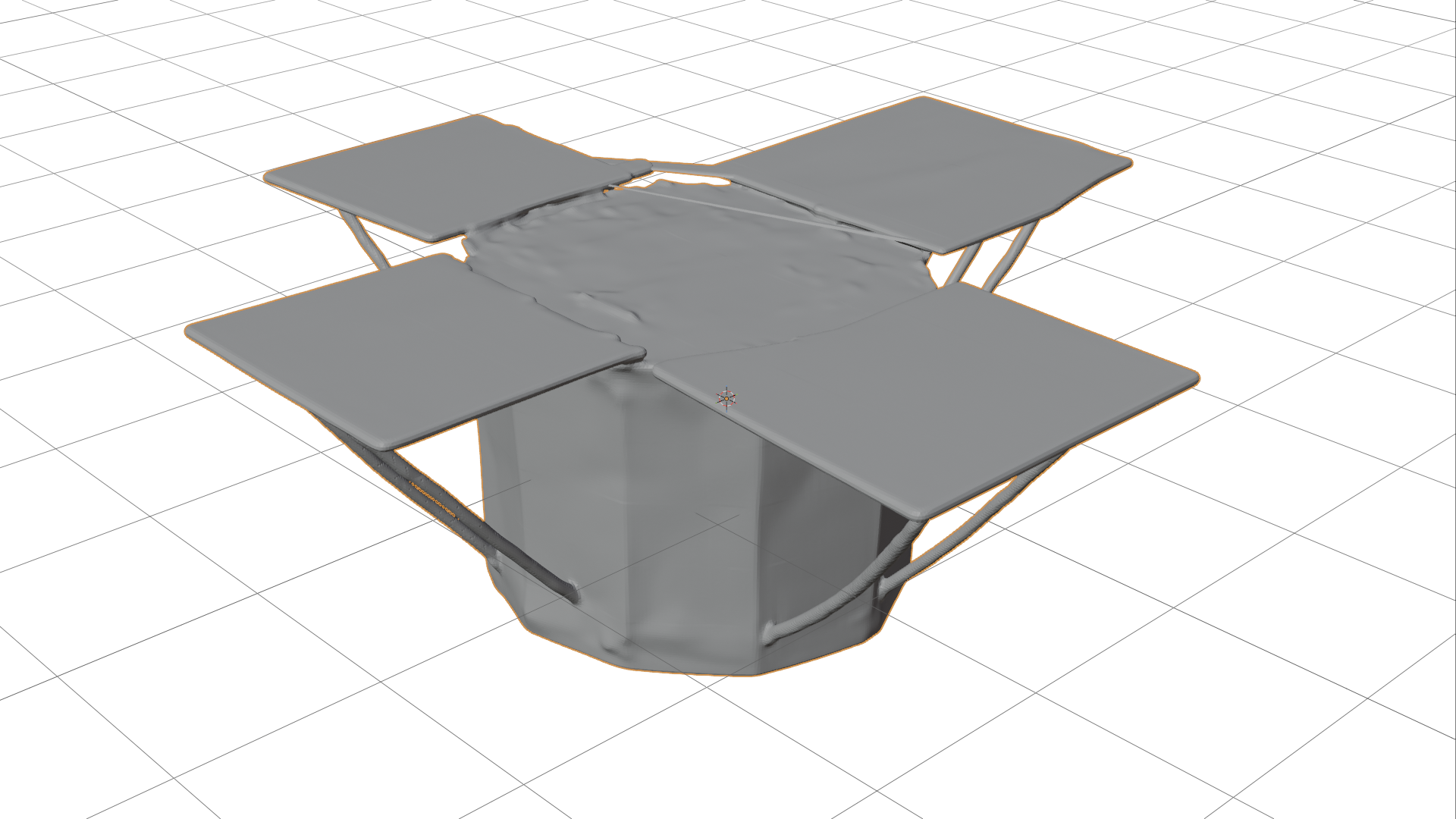} &
        \includegraphics[width=0.19\textwidth,height=1.8cm,keepaspectratio]
        {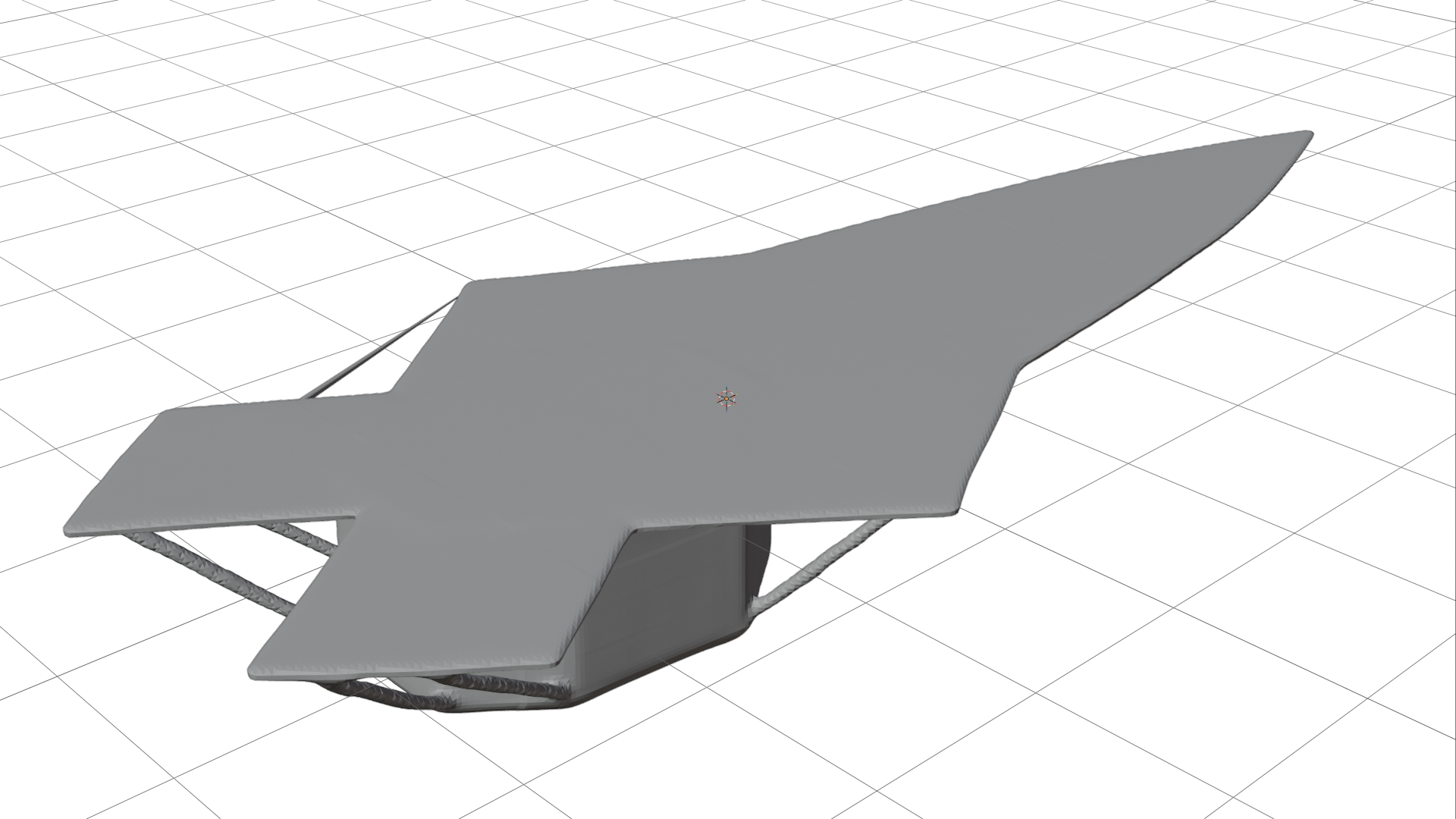} &
        \includegraphics[width=0.19\textwidth,height=1.8cm,keepaspectratio]
        {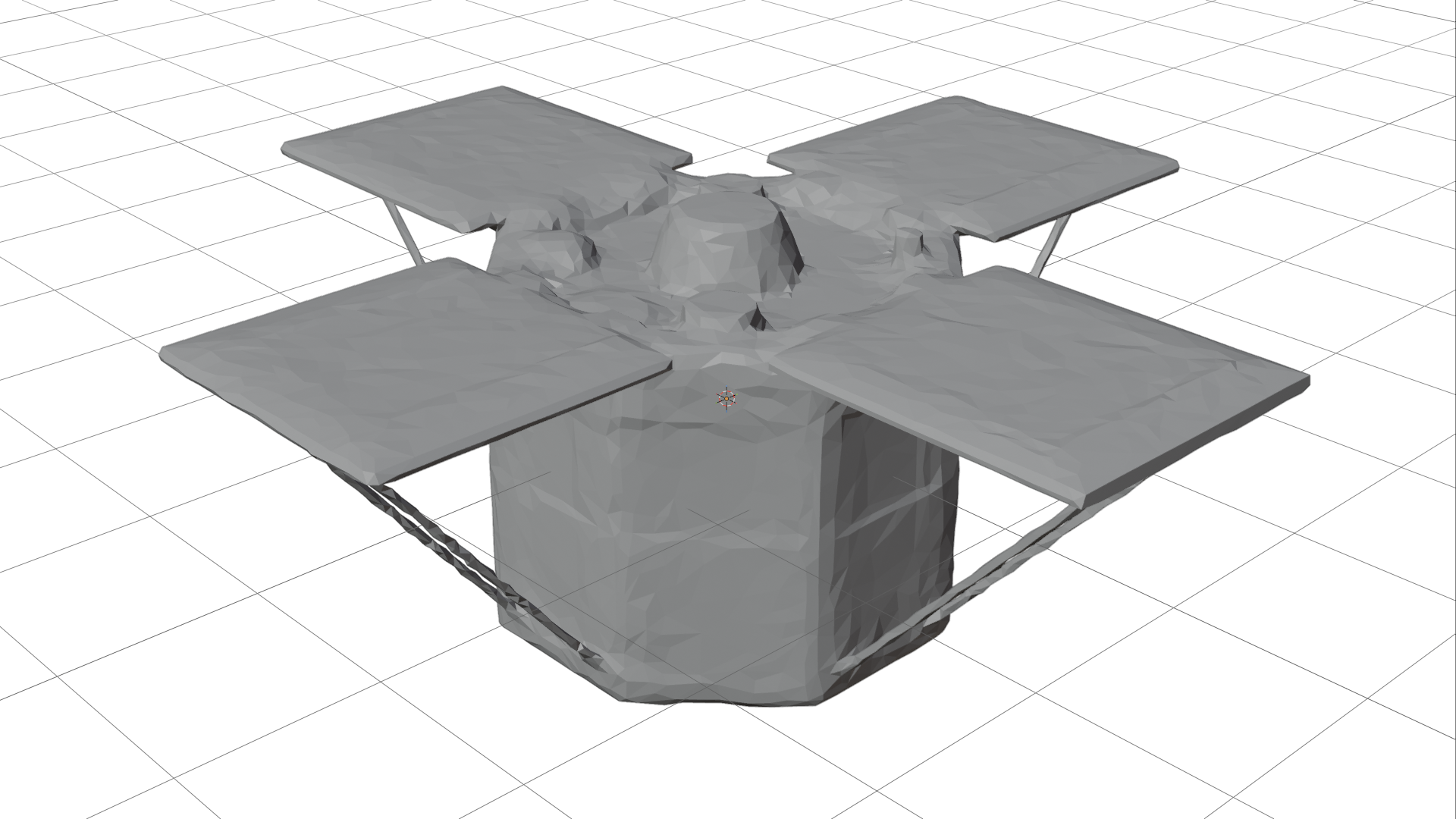}
        \\[2mm]

        &
        \textbf{Input Images} &
        \textbf{Full Model} &
        \textbf{Mini-Turbo} &
        \textbf{Ground Truth}

    \end{tabular}

    \caption{Comparison of simulation input images, Hunyuan3D-2.0 Full
    reconstructions, Hunyuan3D-2.0 Mini-Turbo reconstructions, and
    corresponding ground-truth models.}
    \label{fig:reconstruction-comparison}
\end{figure}

Inspection of the reconstructed meshes revealed two geometric failure modes associated with poor pose-estimation performance. First, reconstructions of long, slender spacecraft, particularly the Mini-Turbo reconstructions of Models 3 and 4, were sensitive to errors in the target’s overall proportions. These discrepancies reduced overlap between the reconstructed geometry and the depth measurements supplied to FP, making the estimated pose ambiguous and unstable.
Second, the reconstruction model did not reliably infer symmetry in unobserved regions. For Model 5, Mini-Turbo introduced unsupported geometry on the occluded side of the spacecraft, increasing the reconstructed volume. The resulting mismatch with the measured depth data led to poor alignment and rapid divergence of the pose estimate.

\subsection{Simulation Experiment: Mesh Resolution}

\begin{figure}[H]
    \centering
    \includegraphics[width=0.8\linewidth]{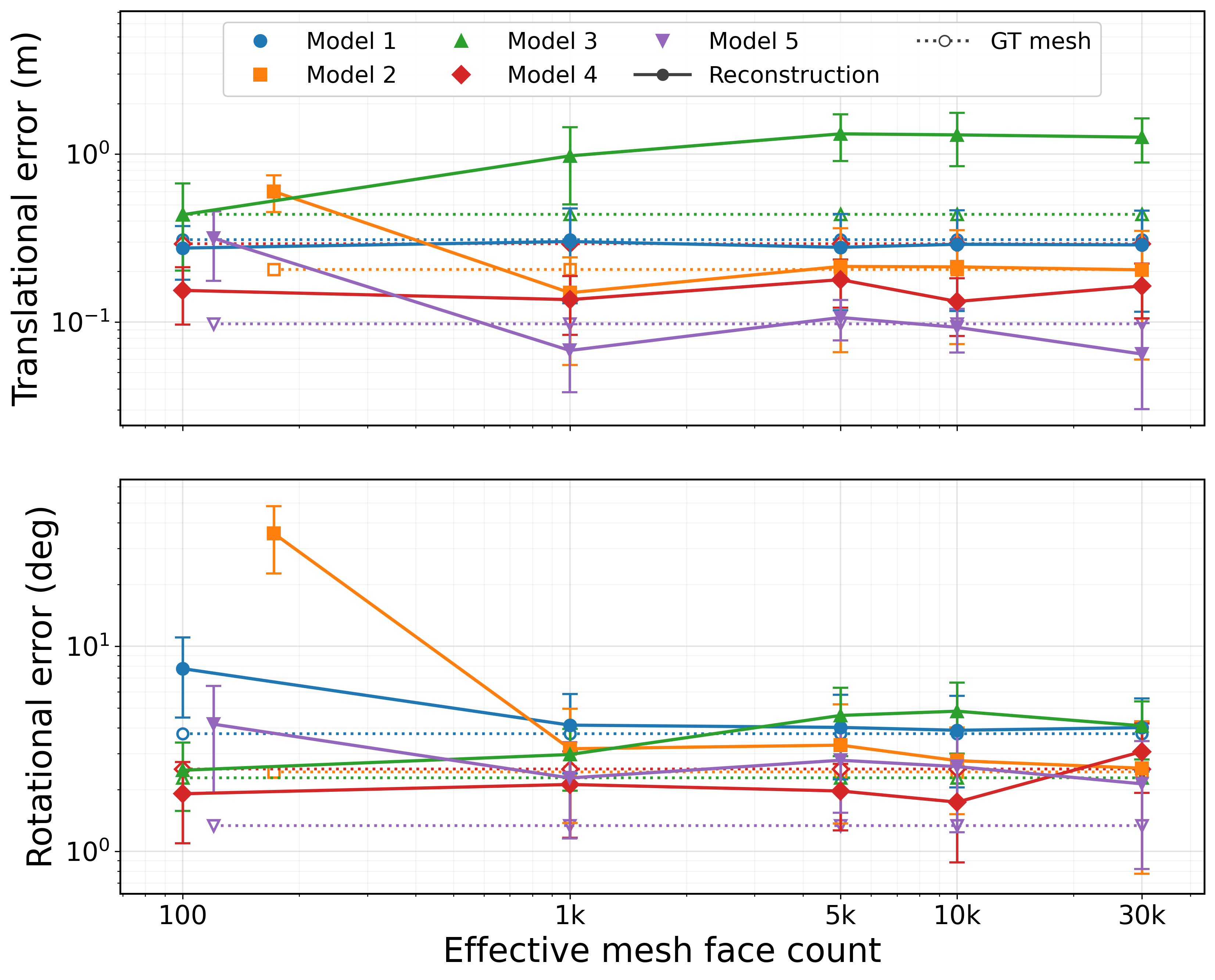}
    \caption{Mean translational and rotational pose-estimation errors as functions
    of reconstructed mesh face count.}
    \label{fig:rq2_mesh_quality}
\end{figure}

The effect of reconstructed mesh resolution on pose-estimation accuracy is evaluated using five spacecraft models. For each spacecraft, the selected full-model reconstruction is decimated to nominal face-count targets of 50,000, 10,000, 5,000, 1,000, and 100 faces. FP is then used to evaluate pose estimates over the same 150-frame orbital segment using each reconstructed mesh. The native GT mesh is also evaluated as a baseline. Because the GT meshes are not decimated, their face counts varied between 4,735 and 58,126. Table~\ref{tab:rq2_results} summarizes the results across all five spacecraft. Each reported value is the mean of the five spacecraft-level tracking errors, while the accompanying standard deviation describes variation between spacecraft models. The temporal variation within each individual test is shown by the error bars in Fig.~\ref{fig:rq2_mesh_quality}. Dotted lines show the corresponding GT-mesh baselines.
\begin{table}[!htbp]
    \centering
    \caption{Across-model pose error by mesh resolution. Values are mean
    $\pm$ sample standard deviation; the GT mesh retains its native resolution.}
    \label{tab:rq2_results}

    \setlength{\tabcolsep}{3pt}
    \renewcommand{\arraystretch}{0.95}

    \begin{tabular}{@{}lcccccc@{}}
        \toprule \midrule 
        Metric & GT & 50k & 10k & 5k & 1k & 100 \\
        \midrule
        $\bar{e}_t \pm s_{\mathrm{model}}$ (m)
        & $0.268 \pm 0.127$
        & $0.396 \pm 0.490$
        & $0.406 \pm 0.507$
        & $0.420 \pm 0.507$
        & $0.326 \pm 0.373$
        & $0.356 \pm 0.169$ \\

        $\bar{e}_r \pm s_{\mathrm{model}}$ (\degree)
        & $2.46 \pm 0.86$
        & $3.17 \pm 0.87$
        & $3.16 \pm 1.20$
        & $3.33 \pm 1.03$
        & $2.93 \pm 0.80$
        & $10.36 \pm 14.22$ \\
        \midrule \bottomrule
    \end{tabular}
\end{table}

The results do not indicate a simple proportional relationship between face count and pose-estimation accuracy. The aggregate performance is relatively similar between the 1k- and 50k-face conditions, with mean translational errors between 0.326-m and 0.420~m and mean rotational errors between $2.93\degree$ and $3.33\degree$. Moreover, the optimal resolution varies between spacecraft models. The 50k condition produced the lowest reconstructed-mesh error for spacecraft model 5, whereas other spacecraft achieve their lowest errors at 100, 1,000, or 10,000 faces. Increasing the face count therefore does not consistently improve tracking accuracy.
An aggressive decimation is less reliable. At the nominal 100-face setting, Models 1 and 2 produce mean rotational errors of $7.77\degree$ and $35.47\degree$, respectively. These failures increase the five-model average to $10.36 \pm 14.22\degree$, despite the condition performing substantially better for several other spacecraft. The large between-model deviation demonstrates that sensitivity to decimation depends strongly on target geometry.
On average, the native GT meshes produced lower translational and rotational errors than any reconstructed-mesh condition, despite often containing fewer faces than the 50k reconstructions. This result indicates that geometric fidelity is more important than polygon count alone. Increasing the resolution of a reconstruction cannot correct inaccurate surfaces, missing geometry, topological artifacts, or geometry inferred incorrectly from the single input view.
The 50k variant is retained for the subsequent range experiment as a standardized, high-resolution operating point. This choice preserves as much reconstructed geometry as permitted by the pose-estimation pipeline and avoids the severe failures observed under aggressive decimation; it should not, however, be interpreted as the optimal resolution for every
spacecraft.



\subsection{Simulation Experiment: Orbital Range}

Orbital-range performance are evaluated using five spacecraft models, each scaled such that its greatest dimension has a real-scale length \qty{2}{m}. In reality, each spacraft would have differently sized greatest dimensions, but to isolate for spacecraft geometry, each model has be scaled to a standard size. For each spacecraft, the RQ1-selected Hunyuan3D-2.0 reconstruction is normalized, decimated to 50,000 faces, and used at every range. The WFOV camera is evaluated from \qty{5}{m} to \qty{50}{m}, while the NFOV camera is evaluated from \qty{100}{m} to \qty{250}{m}. Both reconstructed and GT meshes are tested over the same 150-frame orbital segment. Table~\ref{tab:rq3_results} reports the mean and between-spacecraft standard deviation of the five spacecraft-level results.
\begin{table}[H]
\centering
\caption{Pose-estimation error across spacecraft models at each orbital range.
Values are the mean and sample standard deviation of the five
spacecraft-level mean errors.}
\label{tab:rq3_results}
\setlength{\tabcolsep}{4pt}
\begin{tabular}{lcrrrr}
\toprule \midrule
& & \multicolumn{2}{c}{GT mesh}
& \multicolumn{2}{c}{Reconstructed mesh} \\
\cmidrule(lr){3-4}\cmidrule(lr){5-6}
Range & Camera
& $\bar{e}_t \pm s_{\mathrm{model}}$ (m)
& $\bar{e}_r \pm s_{\mathrm{model}}$ (\si{\degree})
& $\bar{e}_t \pm s_{\mathrm{model}}$ (m)
& $\bar{e}_r \pm s_{\mathrm{model}}$ (\si{\degree}) \\
\midrule
\qty{5}{m}   & WFOV & $0.022 \pm 0.009$ & $1.53 \pm 0.52$
                       & $0.020 \pm 0.010$ & $2.40 \pm 0.97$ \\
\qty{10}{m}  & WFOV & $0.021 \pm 0.008$ & $1.78 \pm 0.41$
                       & $0.029 \pm 0.026$ & $3.11 \pm 2.61$ \\
\qty{25}{m}  & WFOV & $0.023 \pm 0.010$ & $1.79 \pm 0.30$
                       & $0.030 \pm 0.024$ & $3.48 \pm 2.96$ \\
\qty{50}{m}  & WFOV & $0.033 \pm 0.018$ & $3.07 \pm 0.78$
                       & $0.044 \pm 0.024$ & $5.97 \pm 3.52$ \\
\midrule
\qty{100}{m} & NFOV & $0.024 \pm 0.007$ & $1.66 \pm 0.28$
                       & $0.069 \pm 0.099$ & $3.17 \pm 1.88$ \\
\qty{150}{m} & NFOV & $0.029 \pm 0.014$ & $1.87 \pm 0.49$
                       & $0.026 \pm 0.007$ & $3.22 \pm 1.91$ \\
\qty{200}{m} & NFOV & $0.044 \pm 0.029$ & $2.52 \pm 1.37$
                       & $0.039 \pm 0.018$ & $4.19 \pm 2.45$ \\
\qty{250}{m} & NFOV & $1.549 \pm 3.387$ & $12.36 \pm 22.18$
                       & $2.851 \pm 6.273$ & $22.45 \pm 26.61$ \\
\midrule \bottomrule
\end{tabular}
\end{table}

\begin{figure}[H]
\centering
\includegraphics[width=0.6\linewidth]{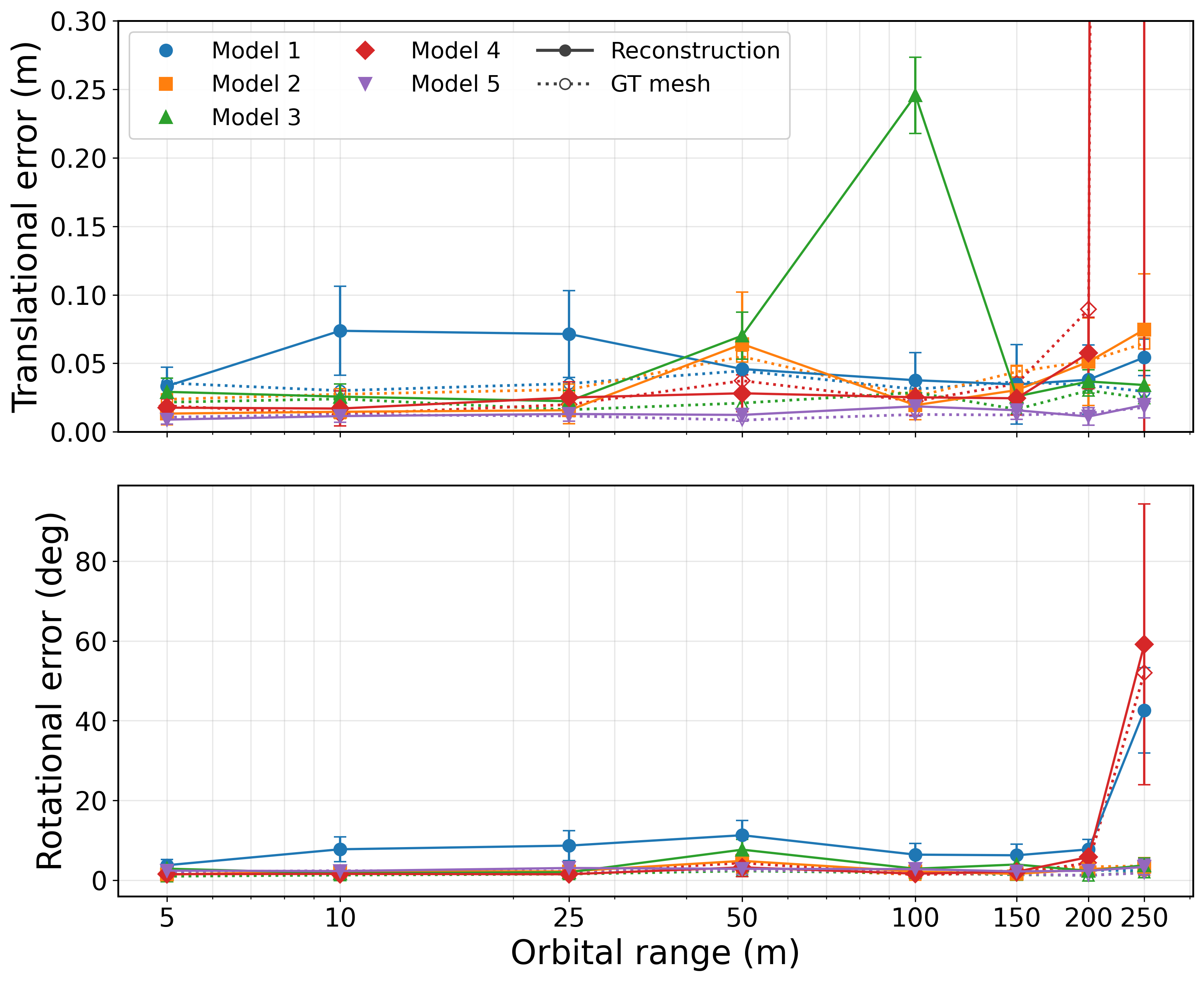}
\caption{Mean translational and rotational pose estimation errors across orbital
range for five spacecraft models. Error
bars show temporal standard deviation, and dotted lines denote GT-mesh
baselines. The translation axis is limited to \qty{0.30}{m} for clarity;
Model~4's \qty{250}{m} results lie outside the displayed range.}
\label{fig:rq3_range_sweep}
\end{figure}

Figure \ref{fig:rq3_range_sweep} shows mean translational and rotational pose estimation errors across orbital range for the five spacecraft models. Error bars denote temporal standard deviation, and dotted lines denote GT-mesh baselines. Within the WFOV regime, reconstructed-mesh error generally increased with range. From \qty{5}{m} to \qty{50}{m}, mean translational error increased from \qty{0.020}{m} to \qty{0.044}{m}, while mean rotational error increased from $2.40\degree$ to $5.97\degree$.
Performance improved at the transition from \qty{50}{m} WFOV to \qty{100}{m} NFOV. Median reconstructed-mesh translation and rotation errors decrease by 45.0\% and 42.8\%, respectively. Although the physical range doubled, the longer NFOV focal length produced an approximately 2.37-times-larger projected target. This suggests that apparent target size may influence pose accuracy more strongly than range alone. However, because range and camera FOV changed simultaneously, their effects cannot be isolated conclusively.
The mean translation error at \qty{100}{m} is elevated by Model~3, which develops a persistent \qty{0.23}{m}--\qty{0.28}{m} translational offset while maintaining moderate rotational error. Excluding this outlier, mean reconstructed-mesh translation error decreased from \qty{0.038}{m} at \qty{50}{m} to \qty{0.025}{m} at \qty{100}{m}.
Reliability declines most noticeably at \qty{250}{m}. Model~4 diverged with both mesh types, while Model~1 experienced a reconstructed-mesh rotational failure. Consequently, the mean errors are substantially larger than the median reconstructed-mesh errors of \qty{0.054}{m} and $3.74\degree$. Thus, the longest range does not cause overall failure, but it increases the likelihood of severe tracking errors.
In general, the pose error does not vary monotonically with physical range. Performance instead depends on the combined effects of range, camera intrinsics, apparent target size, spacecraft geometry, and mesh accuracy.

\subsection{Simulation Experiment: Illumination Condition}

Pose-estimation performance is evaluated over five independently initialized 150-frame intervals spanning different Sun phase angles. Both the best-performing reconstruction selected in RQ1 and the corresponding GT mesh are evaluated. Table~\ref{tab:illumination-results} reports the mean and sample standard deviation across the five spacecraft-level mean errors. 

\begin{table}[H]
    \centering
    \caption{Mean pose-estimation error across the five spacecraft for each illumination region. Bold values indicate the lowest across-region mean for each mesh and error metric.}
    \label{tab:illumination-results}
    \begin{tabular}{@{}llcccc@{}}
        \toprule \midrule
        & & \multicolumn{2}{c}{GT mesh}
        & \multicolumn{2}{c}{Reconstructed mesh} \\
        \cmidrule(lr){3-4}
        \cmidrule(l){5-6}
        Region
        & Oriented phase interval
        & $\bar{e}_t \pm \sigma_{\mathrm{sc}}$ (m)
        & $\bar{e}_r \pm \sigma_{\mathrm{sc}}$ (\degree)
        & $\bar{e}_t \pm \sigma_{\mathrm{sc}}$ (m)
        & $\bar{e}_r \pm \sigma_{\mathrm{sc}}$ (\degree) \\
        \midrule
        Region 1
        & \ang{211.6}--\ang{257.7}
        & $\mathbf{0.201 \pm 0.123}$
        & $\mathbf{1.36 \pm 0.31}$
        & $0.619 \pm 0.753$
        & $7.34 \pm 10.73$ \\

        Region 2
        & \ang{292.2}--\ang{337.3}
        & $0.260 \pm 0.112$
        & $2.47 \pm 0.86$
        & $\mathbf{0.192 \pm 0.078}$
        & $\mathbf{3.52 \pm 0.85}$ \\

        Region 3
        & \ang{356.0}--\ang{360}/\ang{0}--\ang{23.8}
        & $1.261 \pm 2.447$
        & $17.38 \pm 32.79$
        & $0.526 \pm 0.854$
        & $4.71 \pm 2.92$ \\

        Region 4
        & \ang{44.2}--\ang{89.9}
        & $0.381 \pm 0.398$
        & $18.66 \pm 34.46$
        & $2.814 \pm 5.611$
        & $14.58 \pm 20.89$ \\

        Region 5
        & \ang{165.8}--\ang{194.4}
        & $0.843 \pm 1.287$
        & $21.97 \pm 33.25$
        & $1.433 \pm 1.203$
        & $31.05 \pm 33.35$ \\
        \midrule \bottomrule
    \end{tabular}
\end{table}

Region~2 produces the most consistent reconstructed-mesh performance, with across-spacecraft mean translational and rotational errors of $0.192~\mathrm{m}$ and $3.52\degree$, respectively. By contrast, Region~5, which spans the near-opposition geometry around \ang{180}, produces the broadest increase in reconstructed-mesh rotational error. Its mean rotational error was $31.05\degree$, approximately 4.1 times the average across Regions~1--4. Region~5 produces the highest rotational error for Models~1--4, although Model~5 remained comparatively stable at $3.42\degree$.
The translational response is less consistent. Region~5 produces the largest translational error for only two spacecraft, while the aggregate maximum occurs in Region~4 because of a substantial Model~2 tracking failure. GT-mesh failures also occur outside Region~5, indicating that illumination alone does not explain the observed behavior.
Because illumination and viewing direction vary simultaneously along the orbit, and each interval is initialized independently, the experiment demonstrates an association between near-opposition geometry and rotational tracking instability rather than an isolated causal effect of illumination.








    




\begin{figure}[H]
    \centering
    \includegraphics[width=0.6\textwidth]{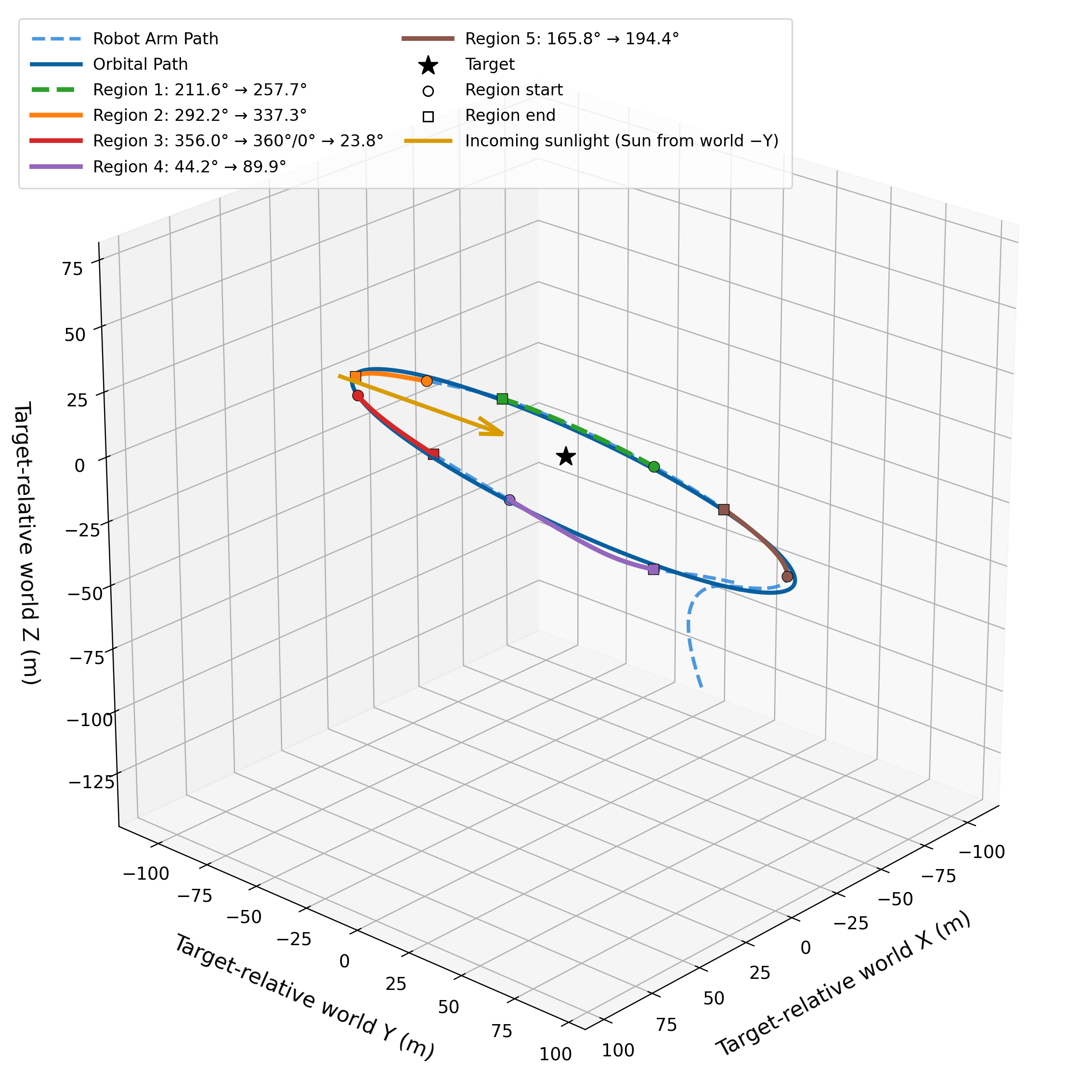}
    \caption{Visualization of Orbit with nominal separation 50 m. Phase angle regions highlighted.}
    \label{fig:orbit_vis}
\end{figure}

In Figure \ref{fig:orbit_vis} above, it should be noted that the recorded arm path is not exactly aligned with the simulated orbital path. To facilitate the translation of simulation environment to hardware, the MuJoCo simulation uses the same xArm7 model used in hardware testing, which is subject to physical joint limits. The physical joint limits result in the slight divergence from the ideal orbital path, particularly as the arm arrives to its initial position because the xArm7 begins to coil about itself. 











\subsection{Hardware Preliminary Experiment}

\begin{figure}[!htbp]
    \centering
    \captionsetup[subfigure]{font=small,skip=2pt}

    \begin{subfigure}[t]{0.24\textwidth}
        \centering
        \includegraphics[
            width=\linewidth,
            height=2.9cm,
            keepaspectratio
        ]{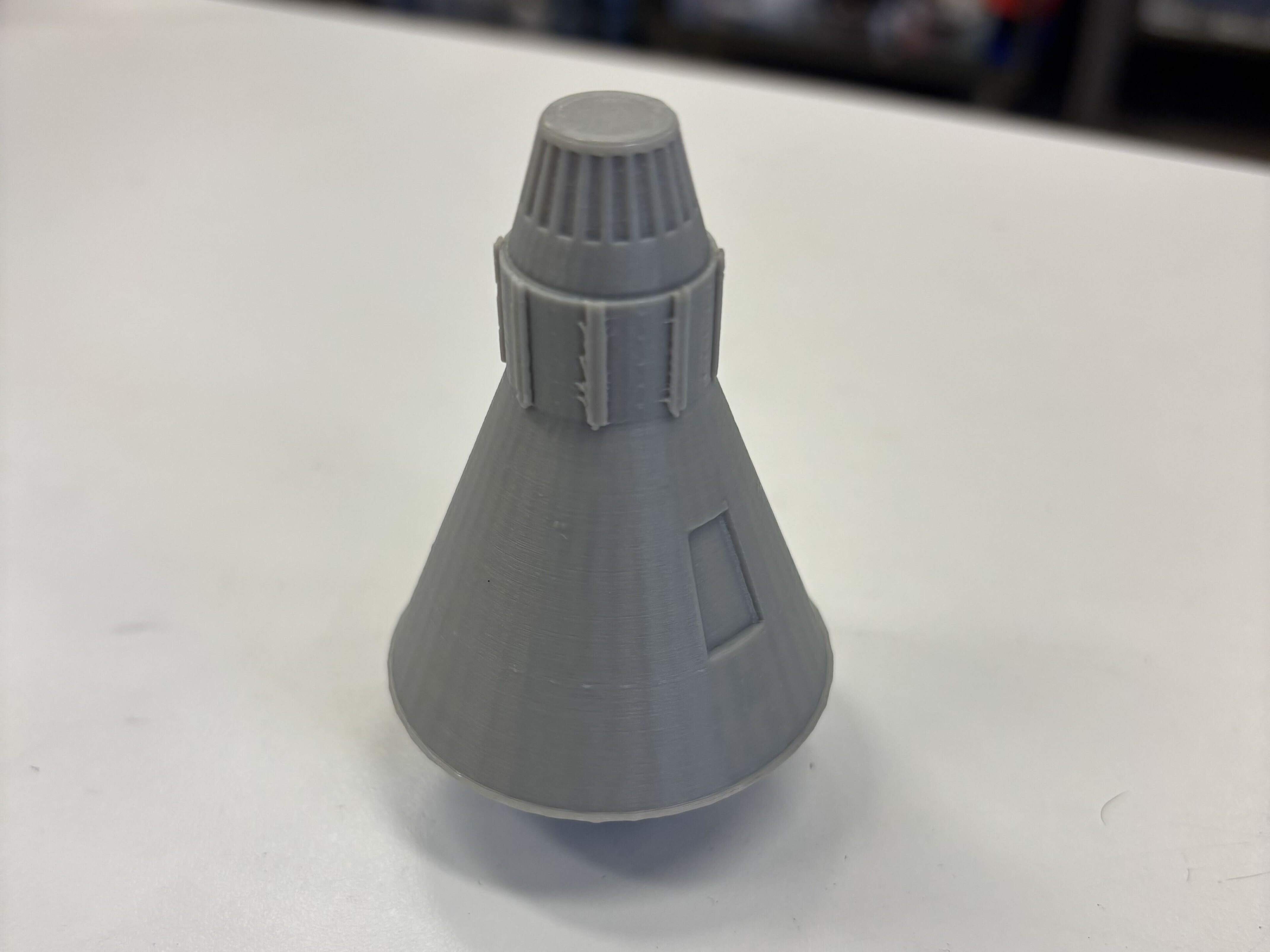}

        \vspace{1mm}

        \includegraphics[
            width=\linewidth,
            height=2.9cm,
            keepaspectratio
        ]{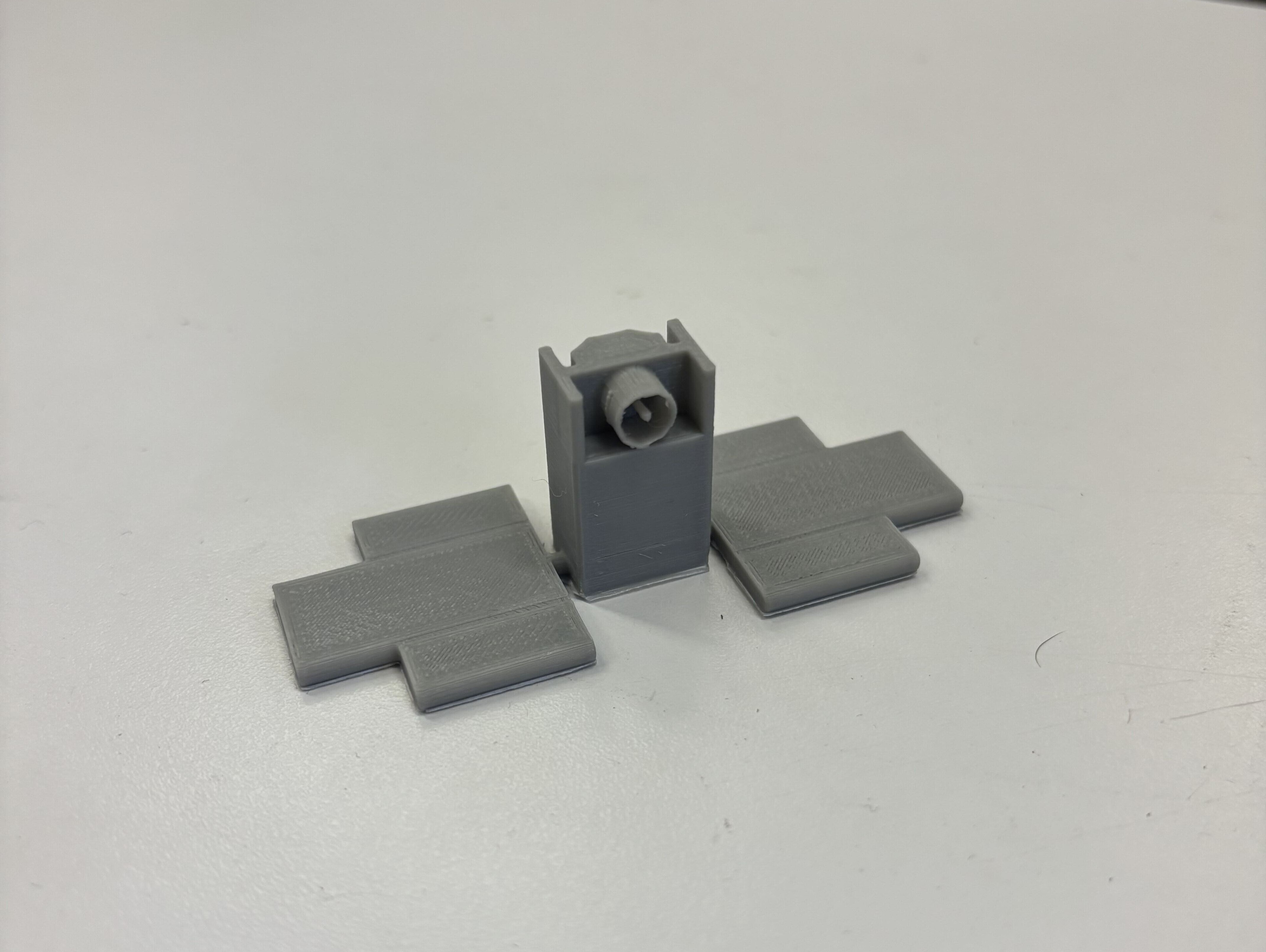}

        \caption{Printed models.}
        \label{fig:left-column}
    \end{subfigure}
    \hfill
    \begin{subfigure}[t]{0.24\textwidth}
        \centering
        \includegraphics[
            width=\linewidth,
            height=2.9cm,
            keepaspectratio
        ]{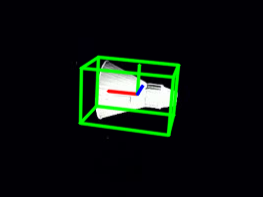}

        \vspace{1mm}

        \includegraphics[
            width=\linewidth,
            height=2.9cm,
            keepaspectratio
        ]{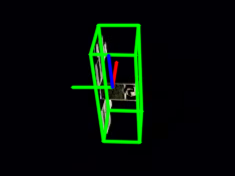}

        \caption{Aligned estimates.}
        \label{fig:center-column}
    \end{subfigure}
    \hfill
    \begin{subfigure}[t]{0.24\textwidth}
        \centering
        \includegraphics[
            width=\linewidth,
            height=2.9cm,
            keepaspectratio
        ]{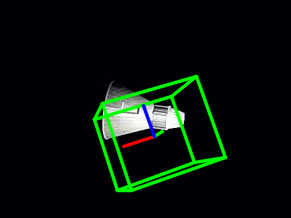}

        \vspace{1mm}

        \includegraphics[
            width=\linewidth,
            height=2.9cm,
            keepaspectratio
        ]{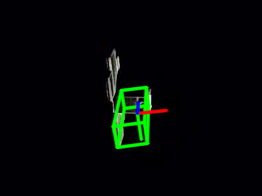}

        \caption{Drifted estimates.}
        \label{fig:right-column}
    \end{subfigure}

    \caption{Representative targets and pose-estimation overlays from the
    preliminary hardware experiment.}
    \label{fig:three-column-comparison}
\end{figure}

The preliminary hardware campaign is conducted to validate the physical test environment, sensing components, and end-to-end reconstruction and pose-tracking pipeline described in Section~\ref{sec:hardware-test-setup}. Because the pose of the physical target is not calibrated in 6-DoF relative to the camera and robot reference frames, translational and rotational pose errors cannot be calculated reliably. The results are therefore evaluated qualitatively by examining recorded RGB sequences overlaid with the pose estimated by FP. Two 3D-printed spacecraft targets with different degrees of geometric symmetry were evaluated under multiple orientations and directional illumination intended to approximate on-orbit lighting. Representative targets and pose overlays are shown in Fig.~\ref{fig:three-column-comparison}.
In most of the examined sequences, the estimated pose remains visually aligned with the target for approximately 10--15~s. The estimate then typically exhibits gradual drift followed by more rapid divergence. More frequent pose re-registration appears to reduce the accumulation of this drift, suggesting that periodic or adaptive re-registration may improve tracking robustness during future hardware experiments.
Pose estimates determined using the nominal GT mesh rather than a reconstruction exhibit a significantly reduced degree of drift, implying that the reconstruction quality is a large determiner of pose accuracy. In the examined sequences, tracking with the GT mesh remains visually aligned over a larger portion of the recorded trajectory. Tracking with a reconstructed mesh is generally most stable when the camera viewpoint remained close to that of the image used for reconstruction. As the observation direction changes, previously occluded or inaccurately reconstructed surfaces become visible, reducing the correspondence between the observed target and the reconstructed geometry and eventually destabilizing the pose estimate. Because the evaluation is qualitative, the respective contributions of reconstruction error, sensor noise, segmentation accuracy, and pose registration could not yet be isolated.
Target symmetry also appears to influence tracking stability. The nearly cylindrically symmetric target exhibits greater apparent rotational ambiguity about its symmetry axis, whereas the target with only a single plane of symmetry generally retains a more stable orientation estimate throughout the recorded sequence. However, only two physical targets are evaluated and differ in characteristics other than symmetry. This observation should therefore be treated as preliminary. A calibrated GT pose and a larger collection of controlled target geometries will be required to quantify the effects of symmetry, reconstruction quality, and re-registration frequency.
Overall, the preliminary campaign successfully demonstrates operation of the complete hardware pipeline using physical RGB and depth measurements under controlled directional illumination. It also identifies pose calibration, reconstruction completeness, target symmetry, and accumulated tracking drift as important subjects for subsequent quantitative hardware evaluation.

\section{Conclusions} \label{sec:concl}

This paper presents the development and initial testing of DreamSat-Bench, a modular RPO testbed designed to bridge the gap between AI-driven 3D reconstruction and autonomous spacecraft navigation. By unifying high-fidelity orbital mechanics with robotic physics engines and physical manipulators, the platform provides a seamless pathway for transitioning vision-based algorithms from software simulation to HITL validation. The core significance of this work lies in its ability to facilitate navigation with unprepared or non-cooperative assets, moving away from a reliance on prior CAD models or cooperative markers and toward a framework capable of characterizing unknown objects in real-time.
The experimental campaigns conducted via the SITL pipeline revealed trade-offs between computational latency and navigational precision. We identified that while generative reconstruction models like Hunyuan3D-v2.0 can produce high-fidelity geometry sufficient for precise 6-DoF tracking, there is a non-linear relationship between reconstruction time and accuracy. Faster, lightweight variants like the Mini-Turbo model demonstrated a significant reduction in inference time but lacked the geometric fidelity required for stable pose estimation. Furthermore, our investigation into mesh resolution suggests that the accuracy of inferred topological features is more vital for tracking stability than raw polygon density, as decimated meshes maintained performance as long as the underlying spacecraft geometry remained intact.
Environmental factors, particularly illumination and orbital range, were also shown to have a profound impact on the reliability of the perception pipeline. Our findings highlighted that near-opposition geometries, where the target is positioned between the chaser and the Sun, induce significant rotational instability, likely due to the lack of distinct shadow-based features and saturated lighting conditions. While transitioning between camera fields-of-view helped maintain performance across varying ranges, the study confirms that the robustness of AI-based navigation is heavily dependent on the interplay between sensor configurations and the dynamic orbital environment.
The preliminary HITL campaign qualitatively validated the end-to-end deployment of the software stack on physical robotic hardware. Although current calibration limits prevent quantitative error reporting in the physical enclosure, the experiments successfully identified key challenges for future work, including the rotational ambiguities presented by symmetric targets and the accumulation of tracking drift over extended trajectories. Future iterations of DreamSat-Bench will focus on the integration of the specialized monocular DreamSat-Pose pipeline and the implementation of high-precision optical tracking systems to provide GT pose data for hardware validation. Ultimately, DreamSat-Bench establishes a standardized benchmarking environment that enables the rapid maturation of the autonomous vision systems necessary for future active debris removal and on-orbit servicing missions.

\section*{Code Availability}
The code for this work can be accessed on GitHub (https://github.com/ARCLab-MIT-X/beavr-bench-dreamsat.git).

\section*{Competing Interests}
The authors declare no competing interests.

\section*{Acknowledgments}
From the MIT side, the research was sponsored by the Department of the Air Force Artificial Intelligence Accelerator and was accomplished under Cooperative Agreement Number FA8750-19-2-1000. The views and conclusions contained in this document are those of the authors and should not be interpreted as representing the official policies, either expressed or implied, of the Department of the Air Force or the U.S. Government. The U.S. Government is authorized to reproduce and distribute reprints for Government purposes notwithstanding any copyright notation herein.

This work has been partially supported by the Spanish Agencia Estatal de Investigacion (AEI) under grant  PID2024-161963OB-C22 (ACTIVATION project). This work has also been supported by the Madrid Government (Comunidad de Madrid-Spain) under the Multiannual Agreement 2023-2026 with Universidad Politécnica de Madrid in the Line A, Emerging PhD researchers.

\section{References}

    \bibliographystyle{unsrt}
    
\begingroup
    \renewcommand{\section}[2]{}%
    \let\oldthebibliography\thebibliography
    \let\endoldthebibliography\endthebibliography
    \renewenvironment{thebibliography}[1]{%
        \begin{oldthebibliography}{#1}%
        \setlength{\itemsep}{0pt}%
        \setlength{\parsep}{0pt}%
    }%
    {\end{oldthebibliography}}
    
    \bibliography{references}
\endgroup

\newpage

\end{document}